\documentclass{article}

\usepackage[preprint]{neurips_2026}

\usepackage[utf8]{inputenc} 
\usepackage[T1]{fontenc}    
\usepackage{hyperref}       
\usepackage{url}            
\usepackage{booktabs}       
\usepackage{amsfonts}       
\usepackage{nicefrac}       
\usepackage{microtype}      
\usepackage{xcolor}         

\usepackage{wrapfig}
\usepackage{microtype}
\usepackage{graphicx}
\usepackage{subcaption}
\usepackage{booktabs} 
\usepackage{xspace}
\usepackage{colortbl}
\usepackage[table]{xcolor}  
\definecolor{grayrow}{rgb}{0.9,0.9,0.9}
\usepackage{hyperref}
\usepackage{multirow}
\usepackage{stfloats}
\usepackage{amsmath}
\usepackage{amssymb}
\usepackage{mathtools}
\usepackage{amsthm}
\usepackage{placeins}

\usepackage[capitalize,noabbrev]{cleveref}
\usepackage[textsize=tiny]{todonotes}
\newcommand\para[1]{{\vspace{5pt} \bf \noindent #1 \hspace{3pt}}}
\newenvironment{packed_itemize}{
	\begin{list}{\labelitemi}{\leftmargin=1.em}
		\setlength{\itemsep}{1pt}
		\setlength{\parskip}{0pt}
		\setlength{\parsep}{0pt}
		\setlength{\headsep}{0pt}
		\setlength{\topskip}{0pt}
		\setlength{\topmargin}{0pt}
		\setlength{\topsep}{0pt}
		\setlength{\partopsep}{0pt}
	}{\end{list}}
\newcommand{\name}{VCR\xspace}

\newcommand{\qs}[1]{\textcolor{black}{#1}}
\newcommand{\jw}[1]{\textcolor{black}{#1}}

\newcommand{\qedit}[1]{\textcolor{black}{#1}}
\newcommand{\rev}[1]{\textcolor{black}{#1}}

\title{VCR: Learning Valid Contextual \\ Representation \rev{for} Incomplete Wearable  Signals}

\author{%
  Yuxuan Weng, Wenhan Luo and Qijia Shao\\
  The Hong Kong University of Science and Technology (HKUST)\\
  \texttt{ywengac@connect.ust.hk, whluo@ust.hk, qijiashao@ust.hk} \\
}

\begin{document}

\maketitle

\vspace{-0.3cm}
\begin{abstract}
  Wearable devices enable continuous health monitoring from multimodal signals, but real-world deployment is hindered by limited labeled data and pervasive sensor incompleteness. While large-scale self-supervised pretraining reduces label dependence, most existing methods assume full modality availability. Current approaches for handling modality missingness often reconstruct entire absent signals, which can encourage hallucinating modality-specific details that are not inferable from the observed sensor signals and degrade robustness. 
  \rev{We propose \name, a self-supervised framework that learns to extract valid representations robust to modality missingness.}
  \name employs an orthogonal tokenizer to enforce strict orthogonal disentanglement by \jw{rectifying} latent manifolds 
  and applying a geometric projection, separating each modality into shared semantics and modality-specific residuals. This design preserves complete information integrity while serving as a structural foundation for robust learning under modality missingness. The resulting tokens are processed by a missing-aware mixture-of-experts backbone that adapts to varying patterns of modality availability. By constraining the objective to reconstruct only the \rev{inferable} shared components of missing modalities, \name effectively mitigates hallucinations of non-inferable modality-specific details. 
  Across multiple health monitoring tasks, \name consistently improves performance and robustness under \qedit{full, single-missing, and multiple-missing modality settings} compared with strong \qedit{supervised and self-supervised baselines.}
\end{abstract}
\setcitestyle{numbers,square}

\bibliographystyle{unsrtnat}

\section{Introduction}
\label{sec:intro}
Wearable devices enable continuous and non-invasive monitoring by capturing synchronized motion and physiological signals such as photoplethysmography (PPG), electrodermal activity (EDA), accelerometer (ACC), and skin temperature (TEMP). Beyond basic signal processing (e.g., estimating heart rate from PPG), these modalities support a broad range of downstream health applications, including emotion recognition~\cite{schmidt2018introducing, bari2020automated}, sleep staging~\cite{zhai2020making,koushik2018real}, and activity recognition~\cite{sheng2020weakly, liu2020giobalfusion}. 
However, reliable generalization in real-world wearable deployments remains challenging due to two coupled factors:  (i) label scarcity across diverse tasks~\cite{erturk2025beyondsensor}, and (ii) pervasive modality missingness
\footnote{\rev{In this paper, ``missingness" refers to the complete absence of a modality throughout the given window.}} caused by hardware constraints and user-dependent effects~\cite{narayanswamy2025scaling} (e.g., some wearables omit EDA/TEMP entirely, users disabling PPG to save power, and motion/firmware issues can cause ACC streams to drop out).

%
\begin{figure}[t]
    \centering 
    \hspace*{-5ex}
    \vspace{1ex}
    \setlength{\abovecaptionskip}{1pt}  
    \includegraphics[width=\linewidth, trim=0.4cm 0.9cm 0cm 0.95cm, clip]{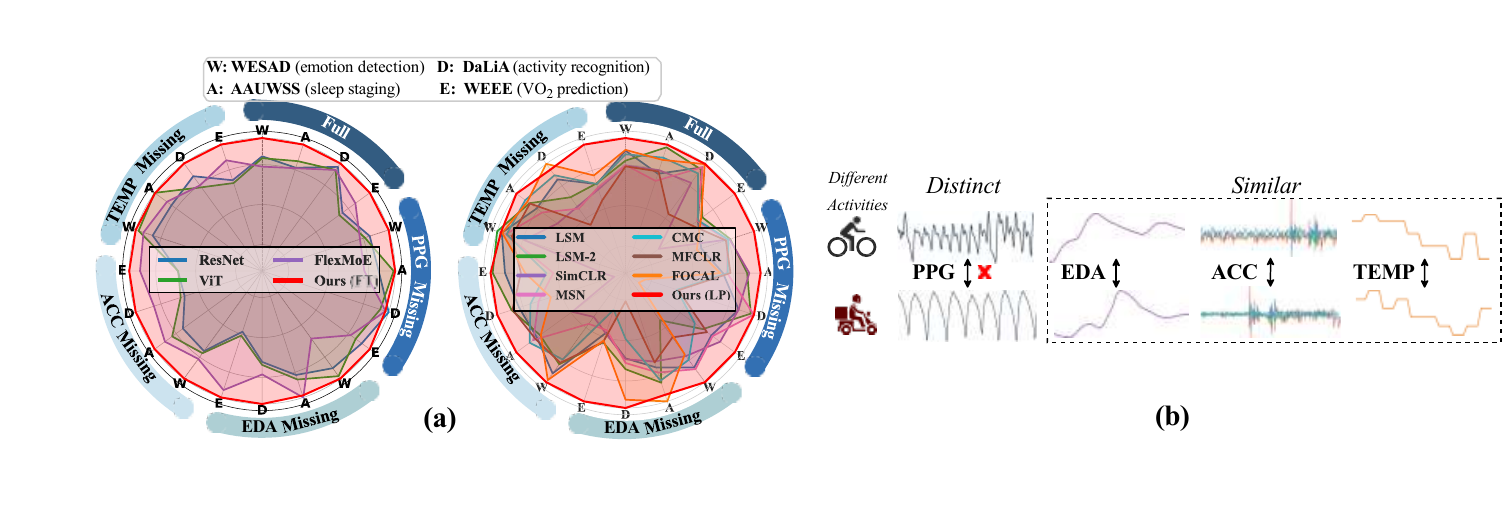}
    \vspace{-0.3cm}
    \caption{
    \rev{(a) \textbf{Performance under full and single-modal missingness} (see Experiments for multiple missingness).}
    We report normalized Macro-F1 \rev{(value/max) and MAE (min/value)} within each dataset to compare methods across tasks and datasets. 
    Left: comparison to supervised baselines trained on labeled data only. Right: comparison to pretrained/self-supervised baselines. Higher is better.
    (b) \qedit{\textbf{Modality-specific information can be non-identifiable from observed sensors.}  Two activity windows may exhibit similar EDA, ACC, and TEMP signals while showing distinct PPG morphology. } 
    }
    
    \label{fig:radar} 
    \vspace{-1cm}
\end{figure}

Recent progress in self-supervised learning and foundation models~\cite{xurelcon,pillaipapagei,zhang2025sensorlm} has made label-efficient wearable intelligence increasingly feasible: models can be pretrained on large-scale unlabeled data streams and transferred to downstream tasks with limited labeled data~\cite{narayanswamy2025scaling,erturk2025beyondsensor}. 
However, most existing models implicitly assume full modality availability during pretraining, thereby limiting their real-world utility 
\qedit{under heterogeneous modality missingness.}
While contrastive learning approaches~\cite{tian2020contrastive,assran2022msn,zhang2022self} can handle missing modalities by aligning shared representations, 
\qedit{their emphasis on cross-modal commonality may suppress modality-specific cues}. 
\jw{Conversely, missingness-aware reconstruction}~\cite{xu2025lsm2} often relies on masked reconstruction of the entire missing signal. 
\qedit{When the absent modalities contain modality-specific details that are not identifiable from the observed sensors, such objectives can encourage hallucination-like completions and inject spurious information into the learned representation.}
Meanwhile, task-specific modality-routing methods~\cite{yun2024flexmoe,han2024fusemoe} may improve a particular downstream task but are not naturally reusable across multiple tasks. 
\qedit{These limitations motivate a general-purpose pretrain model that remains valid and robust under train-on-full, test-on-missing scenarios. Achieving this goal requires addressing the following technical challenges.} 
\vspace{-0.5cm}

\begin{packed_itemize}
    \item \textbf{Non-identifiability of Missing Modality-Specific Information.} When a modality is absent, its modality-specific details cannot be reliably inferred from the remaining sensors as illustrated in Figure~\ref{fig:radar}(b). Nevertheless, reconstruction-based objectives can still incentivize generating them, leading to hallucination-like completions and injecting noise into representations.
    
    \item \textbf{Availability-Dependent Cross-Modal Dependencies.} Different missing-modality patterns
    \rev{determine} which inter-modal correlations are observable, making it challenging for a 
    \rev{unified} dense architecture to 
    \rev{model all modality combinations} without interference. 
    
    \item \textbf{Information Loss Caused by Imbalanced Learning.} Wearable modalities differ substantially in sampling rate, variance, and semantic content. As a result, masked reconstruction can be dominated by high-variance modalities and underutilize other modalities, while contrastive alignment can overemphasize shared semantics and underrepresent modality-specific cues. 
    
\end{packed_itemize}
\vspace{-0.2cm}

\qedit{To address these obstacles, we propose \name, a missingness-robust general-purpose framework for multimodal wearable sensor data, pre-trained on raw PPG, EDA, ACC, TEMP signals. 
\name consists of two stages: an Orthogonal Tokenizer that separates each modality into shared and specific components, and a Missing-Aware MoE Backbone that learns contextual representations under diverse modality-availability patterns.
Together, these components enable \name to preserve information under full observations while avoiding unsupported reconstruction when modalities are missing.}

Our key contributions are as follows:
\vspace{-0.2cm}
\begin{packed_itemize}

    
    
    \item \rev{\textbf{Validity-Aware Learning:}} 
    \qedit{
    To address the non-identifiability of missing modality-specific information, we propose an orthogonal tokenizer that rectifies encoder representations, applies whitening, and uses a learned orthogonal projection to decompose each modality into modality-shared semantics and modality-specific residuals. During structural missingness,  our validity-aware objective reconstructs only inferable shared components of absent modalities, thereby avoiding unsupported modality-specific reconstruction and mitigating hallucination-like completions.} 
    

    \item \textbf{Missing-Aware MoE Backbone:} 
    \qedit{To handle availability-dependent cross-modal dependencies, we introduce a sparse MoE backbone with learnable partition embeddings. This preserves modality/component identity and enables experts to specialize under different observed modality sets.}
    
    \item \rev{\textbf{Two-Stage Information Preservation:} 
    To mitigate imbalanced and conflicting SSL objectives,  we first train the tokenizer to preserve shared semantics via cross-modal contrast and specific details via intra-modal reconstruction. }
    \qedit{We then train the backbone with masked latent reconstruction, reducing domination by high-frequency or high-variance raw signals.}

\end{packed_itemize}
\vspace{-0.2cm}

We evaluate \name on a comprehensive benchmark spanning three classification tasks (i.e., emotion detection, activity recognition, sleep staging), \rev{and one regression task (i.e., $\text{VO}_2$ prediction).} 
\rev{Across these tasks as shown in Figure~\ref{fig:radar}(a)}, \name achieves state-of-the-art (SOTA) performance under both full-modality and all missing-modality settings, demonstrating \jw{its effectiveness }
in preserving information integrity and ensuring robustness against modality missingness.
We will make the code publicly available upon acceptance to support reproducibility and future research.
\vspace{-0.3cm}
\section{Method}
\vspace{-0.3cm}
This section introduces VCR. We first formalize the problem and motivate disentanglement (Sec.~\ref{preliminary}), then describe the overall framework (Sec.~\ref{overall}), the orthogonal tokenizer (Sec.~\ref{tokenizer}), and the missing-aware MoE backbone with training objectives (Sec.~\ref{backbone}).

\vspace{-0.2cm}
\subsection{Preliminary}\label{preliminary}
\jw{\textbf{Problem Formalization.}}
\qs{We consider multimodal wearable time-series data collected from $M$ sensing modalities, including PPG,  EDA,  ACC, and TEMP. \jw{After synchronization and \rev{non-overlapping} windowing, each sample is denoted as $X = \{\mathbf{x}^{m}\}_{m}^M$. Here, 
$\mathbf{x}^m \in \mathbb{R}^{T_m \times C_m}$} denotes the sequence for modality $m$, 
where \jw{$T_m$} is \jw{the number of time steps (the product of window duration and sampling rate) and $C_m$ is the number of channels.}
In actual deployments, not all modalities are always available due to heterogeneous device configurations, sensor failures, or user non-compliance. 
\qedit{In this paper, a missing modality refers to the complete absence of an entire modality throughout the current window.}
\jw{We formalize this by defining the observed and missing modality sets as} $\mathcal{M}_{\text{obs}}$ and $\mathcal{M}_{\text{miss}}$, respectively, \jw{where $\mathcal{M}_{\text{obs}}, \mathcal{M}_{\text{miss}} \subseteq \mathcal{M}=\{1,\dots,M\}$. Accordingly, we denote the set of observed modalities as $\mathcal{X}_{obs} = \{\mathbf{x}^k | k \in \mathcal{M}_{obs}\}$ and the set of missing modalities as $\mathcal{X}_{miss} = \{\mathbf{x}^k | k \in \mathcal{M}_{miss}\}$.} } 
\qedit{When handling missingness, the model receives only $\mathcal{X}_{\mathrm{obs}}$ from the current window.} 

\textbf{Information Disentanglement Hypothesis.} We assume the latent information of each modality $\mathbf{x}^{m}$ can be structurally disentangled into two components~\cite{bousmalis2016dsn,huang2025ssole}: a \textbf{modal-shared} component $\mathbf{s}^{m}$ that captures semantics correlated with other modalities 
(e.g., heart-rate-related patterns in PPG may correlate with ACC micro-vibrations), and a  \textbf{modal-specific} component $\mathbf{p}^{m}$ that is unique to $m$ (e.g., skin impedance drift in EDA). By definition, $\mathbf{p}^{m}$ contains the information \qs{that is not \emph{reliably identifiable} from other modalities.} \qs{Accordingly, our framework encourages the disentangled components to be statistically independent, i.e., $I(\mathbf{s}^{m}; \mathbf{p}^{m}) = 0$.}

\textbf{Hallucination under structural modality missingness.} A reconstruction objective is well-posed only when the target information is identifiable from the available context. Under our decomposition, a modality signal is represented by $(\mathbf{s}^m, \mathbf{p}^m)$. Standard masked autoencoding approaches implicitly maximize the likelihood of reconstructing the full modality signal:
\jw{$\log p(\mathbf{x}^m \mid \mathcal{X}_{obs}) \;=\; \log p(\mathbf{s}^m, \mathbf{p}^m \mid \mathcal{X}_{obs})$.}
When modality $m$ is structurally missing, the \jw{modal-specific component } 
$\mathbf{p}^m$ is not identifiable from the observed modalities. 
\rev{Thus, reconstructing the full signal forces the model to generate plausible but ungrounded modality-specific patterns, which we term \emph{hallucination}.}

\vspace{-0.2cm}
\subsection{Overview of the Framework}\label{overall}
\vspace{-0.2cm}

\begin{figure}[t]
    \centering 
    \vspace{1ex}
    \setlength{\abovecaptionskip}{3pt}  
    \includegraphics[width=1\linewidth]{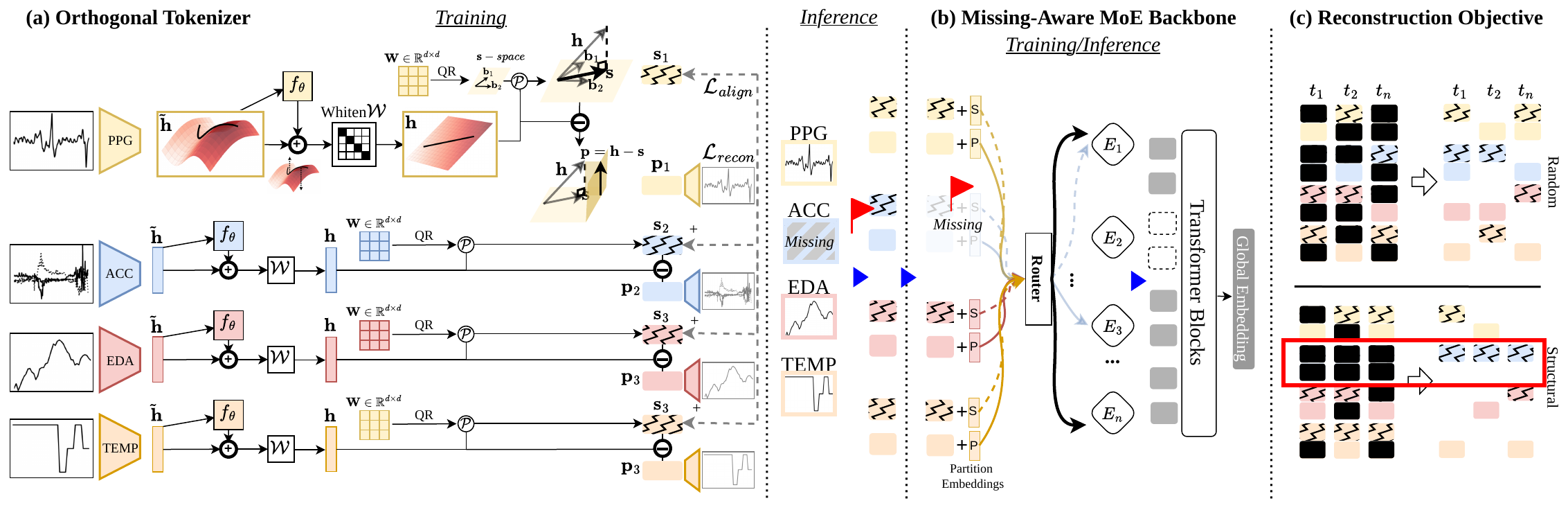}
    \caption{Overview of \name. \textbf{(a) Orthogonal Tokenizer}: \rev{During training,} encoders map raw signals to embeddings $\tilde{\mathbf{h}}$, which are rectified and projected by $\mathcal{P}$ into orthogonal shared $\mathbf{s}$ and specific $\mathbf{p}$ subspaces. Optimization combines cross-modal contrastive learning and intra-modal reconstruction \rev{in a full modality setting}. \rev{During inference, our tokenizer uses a zero-padded placeholder with a ``missingness flag" for missing modality to ensure tensor-shape consistency.}
    \textbf{(b) Missing-Aware MoE Backbone:} Learnable partition embeddings are added to tokens and missing modalities \rev{with flags} are \rev{completely} discarded at the input level. 
    \textbf{(c) Masked Token Reconstruction:} The model performs reconstruction in the latent space. Under structural masking (entire modality missing, highlighted in red), the objective is limited to recovering only the \rev{inferable} shared component $\mathbf{s}$.
    }
    \label{fig:overview} 
    \vspace{-2ex}
\end{figure}

\rev{Figure~\ref{fig:overview} illustrates the proposed two-stage \name framework for learning robust multimodal representations from wearable time-series under heterogeneous modality missingness. }
Given \qedit{a windowed sample}
with an availability pattern, VCR consists of two components: (i) an \emph{Orthogonal Tokenizer} that \qedit{is pretrained on fully observed windows} and decomposes each modality into modality-shared and modality-specific factors, and (ii) a \emph{Missing-Aware MoE Backbone} that \qedit{receives only the tokens from observed modalities} and adapts computation to different observed modality sets.
\qedit{During backbone pretraining, we use both random patch masking and structural masking, where 1--3 entire modalities are removed at the sample level.}

\vspace{-1ex}
\subsection{Orthogonal Tokenizer}\label{tokenizer}
Disentangling modal-shared and modal-specific features is a prerequisite for \jw{explicitly} preserving information integrity\jw{~\cite{bousmalis2016dsn}} and enabling selective reconstruction. Existing disentanglement methods typically impose soft constraints (e.g., orthogonality penalties \qs{in loss functions}) between the shared and specific components~\cite{liu2023focal}.
Such an approach can lead to information leakage between the two components, causing $\mathbf{s}$ to attempt to reconstruct $\mathbf{p}$-related information.

\textbf{Geometric Construction.} We define a learnable parameter matrix $\mathbf{W} \in \mathbb{R}^{d \times d}$, where $d$ denotes the embedding dimension. We then perform QR decomposition on $\mathbf{W}$ during the forward pass: $\mathbf{W} = \mathbf{Q} \mathbf{R}$. After determining the rank $r$ of the modal-shared space, we extract the first $r$ columns of the orthonormal matrix $\mathbf{Q}$ to form a basis matrix $\mathbf{B} \in \mathbb{R}^{d \times r}$, such that $\mathbf{B}^\top \mathbf{B} = \mathbf{I}_r$. Using this basis, we construct an orthogonal projection operator $\mathcal{P} = \mathbf{B} \mathbf{B}^\top \in \mathbb{R}^{d \times d}$. By construction, $\mathcal{P}$ is symmetric ($\mathcal{P}^\top = \mathcal{P}$) and idempotent ($\mathcal{P}^2 = \mathcal{P}$), satisfying the necessary and sufficient conditions for an orthogonal projection operator.

\textbf{\jw{Structural} Disentanglement and Independence.} 
Since the raw latent representations from the encoder may reside on a highly non-linear manifold, directly applying a linear projection $\mathcal{P}$ might be insufficient to disentangle complex dependencies.
To address this, we introduce a 
\jw{lightweight residual adapter $f_\theta$, which predicts an additive correction $f_\theta(\tilde{\mathbf{h}})$, effectively rectifying the representation to better align with the target subspace.}
The rectified and whitened embedding $\mathbf{h}$ is obtained via:
\begin{align}
    \mathbf{h} = \mathcal{W}(\tilde{\mathbf{h}} + f_\theta(\tilde{\mathbf{h}})),
\end{align}
where $\mathcal{W}$ is the whitening process~\cite{zbontar2021barlow} and $\tilde{\mathbf{h}}$ is the original encoder embedding.
Then we employ $\mathcal{P}$ to project $\mathbf{h}$ onto the modal-shared subspace, yielding the shared representation: 
\begin{align}
   \mathbf{s} = \mathcal{P} \mathbf{h}.
\end{align}
Accordingly, the modality-specific representation is defined as the residual: 
\begin{align}
    \mathbf{p} = \mathbf{h} - \mathbf{s}=(\mathbf{I} - \mathcal{P}) \mathbf{h},
\end{align}
which constitutes the strict geometric orthogonal complement of $\mathbf{s}$. Crucially, this geometric orthogonality translates to statistical independence under the condition that the input embeddings are whitened. Assuming the covariance of the whitened input approximates identity ($\text{Cov}(h) \approx \mathbf{I}$), we can derive that $\mathbf{s}$ and $\mathbf{p}$ are uncorrelated: $$\begin{aligned}
\text{Cov}(\mathbf{s}, \mathbf{p}) &= \mathbb{E}[\mathbf{s}\mathbf{p}^\top] = \mathbb{E}[\mathbf{\mathcal{P}}\mathbf{h}\mathbf{h}^\top(\mathbf{I} - \mathbf{\mathcal{P}})^\top] = \mathbf{\mathcal{P}}\mathbb{E}[\mathbf{h}\mathbf{h}^\top](\mathbf{I} - \mathbf{\mathcal{P}}) \approx \mathbf{\mathcal{P}}\mathbf{I}(\mathbf{I} - \mathbf{\mathcal{P}}) = \mathbf{\mathcal{P}} - \mathbf{\mathcal{P}}^2 = \mathbf{0}.
\end{aligned}$$

Under the standard assumption that latent representations follow a multivariate Gaussian distribution, this zero covariance implies statistical independence, i.e., $I(\mathbf{s}; \mathbf{p}) \approx 0$. This guarantees that $\mathbf{s}$ contains no information about $\mathbf{p}$, effectively eliminating feature redundancy by construction.

\textbf{Hybrid Optimization Objectives.}
With the architecture ensuring structural separation, we need to populate $\mathbf{s}$ and $\mathbf{p}$ with semantic correctness and align the shared spaces across modalities. A single objective is insufficient for this dual purpose. Solely relying on masked reconstruction would still force the model to generate specific details using other modalities. Conversely, applying contrastive learning to $\mathbf{p}$ is detrimental, as standard data augmentations (e.g., scaling ACC data or adding noise to TEMP) can destroy the physical semantics of motion and physiological signals.
Therefore, we adopt a hybrid strategy with three specific objectives.

\textit{Cross-Modal Alignment.} To force the shared subspace to capture inter-modal semantics, we apply contrastive learning~\cite{chen2020simclr} exclusively to $\mathbf{s}$. We employ a masked contrastive loss to handle the temporal structure of motion and physiological signals. 
For a given anchor token $\mathbf{s}_i^m$ (from modality $m$ in sample $i$), we define the positive set $\mathcal{S}_i^+$ as the synchronized tokens from other modalities $\{\mathbf{s}_i^n\}_{n \neq m}$, and the full candidate set as $\mathcal{A}_i = \mathcal{S}_i^+ \cup \mathcal{S}_i^-$. 
For brevity, let $\phi(\mathbf{u}, \mathbf{v}) = \exp({sim}(\mathbf{u}, \mathbf{v}) / \tau)$. The loss for modality $m$ is defined as:


\begin{equation}
\mathcal{L}_{align}(\mathbf{s}_i^m)
= \log \sum_{\mathbf{s} \in \mathcal{A}_i} \phi(\mathbf{s}_i^m, \mathbf{s})
- \log \sum_{\mathbf{s}^+ \in \mathcal{S}^+_i} \phi(\mathbf{s}_i^m, \mathbf{s}^+).
\label{eq:align_loss}
\end{equation}
\textit{Intra-Modal Reconstruction.} To ensure the specific component $\mathbf{p}$ captures the unique details of each modality (e.g., high-frequency noise in ACC), we apply a reconstruction objective. A modality-specific decoder $D^m$ attempts to recover the original input $\mathbf{x}^m$ solely from $\mathbf{p}^m$. With $\mathcal{L}_{{recon}}$, $\mathbf{p}$ retains all information not captured by $\mathbf{s}$:
\begin{align}
    \mathcal{L}_{{recon}}(\mathbf{p}^m) = \|\mathbf{x}^m - D^m(\mathbf{p}^m)\|^2_2.
\end{align}

\textit{Whitening Regularization.} To satisfy the independence derivation condition ($\text{Cov}(\mathbf{h}) \approx \mathbf{I}$), we impose a whitening penalty on the encoder outputs. We minimize the squared Frobenius norm of the off-diagonal elements of the covariance matrix of $\mathbf{h}^m$: 
\begin{align}
    \mathcal{L}_{{white}}(\mathbf{h}^m) = \|\text{Cov}(\mathbf{h}^m) - \text{diag}(\text{Cov}(\mathbf{h}^m))\|_F^2.
\end{align}

\textbf{Total Tokenizer Loss.} The final training objective for the tokenizer aggregates these components across all modalities:
\begin{equation}
\begin{split}
\mathcal{L}_{tok} = \sum_{m \in \mathcal{M}} \Big( & \mathcal{L}_{align}(\mathbf{s}^m) + \lambda_{recon}\mathcal{L}_{recon}(\mathbf{p}^m) + \lambda_{white}\mathcal{L}_{white}(\mathbf{h}^m) \Big),
\end{split}
\end{equation}
\rev{where $\lambda_{{recon}}$ and $\lambda_{{white}}$ balance the specific reconstruction and whitening constraints.}

\subsection{Missing-Aware MoE Backbone}\label{backbone}
\textbf{Architecture.} Wearable sensors exhibit intrinsic physical and semantic entanglements. 
For example, interpreting EDA activation often requires temperature signals to distinguish between physical activity and stress. However, when the model learns to handle modality missingness, it tends to weaken cross-modal dependencies. In traditional dense architectures, using a shared set of parameters to learn both objectives leads to modality conflict, preventing the model from effectively capturing cross-modal entanglements.
To address this, we incorporate a MoE~\cite{shazeer2017outrageously} block at the early layer of the backbone. Each expert autonomously learns specialized combinations of modalities. When certain modalities are missing, unaffected experts naturally gain greater influence, ensuring globally optimal representations under the current conditions without compromising for the missing modality patterns. Meanwhile, conventional approaches to modality missingness often rely on zero-padding or special tokens, which introduce noise and degrade model performance. Instead, we discard missing modalities to avoid such interference. 
To ensure the router is aware of each token's modality origin,
we inject learnable partition embeddings that explicitly identify the semantic role (i.e., modality identity and component type $\mathbf{s}$ vs. $\mathbf{p}$) of each token. By inheriting the structural inductive bias from the Orthogonal Tokenizer, this design simplifies the decision boundary for the router, significantly reducing optimization difficulty and allowing the backbone to focus on robust inference.

\textbf{Optimization Objective.} The backbone is tasked with mining deep contextual semantics from the tokenizer outputs and handling diverse missingness patterns. Given the difficulty of augmenting human physiological signals, we employ masked \jw{reconstruction} for self-supervised learning. However, reconstructing the raw signals of missing modalities forces the model to predict modal-specific details that are statistically unpredictable from the remaining context. To avoid this and mitigate modality competition arising from variations in frequency and complexity,  we perform masked reconstruction on the latent embeddings~\cite{baevski2022data2vec,wei2022masked} ($\mathbf{s}$ and $\mathbf{p}$) with two complementary masking strategies as the optimization objectives.

\textit{Random Masking ($\mathcal{L}_{rand}$).} We randomly mask~\cite{he2022masked} a proportion $p_r$ of temporal patches for all available modalities. The model is required to reconstruct both the shared ($\mathbf{s}$) and specific ($\mathbf{p}$) embeddings for these masked patches, as the full information is theoretically recoverable from the context. Let $\mathcal{M}$ denote the set of modalities, and $\mathbb{M}_{t}^m \in \{0, 1\}$ be the binary mask indicator for the patch at time $t$ of modality $m$. The loss is defined as:
\vspace{-0.1cm}
\begin{align}
    \mathcal{L}_{rand} = \sum_{m \in \mathcal{M}} \sum_{t=1}^{T_m} \mathbb{M}_{t}^m \cdot \left( \| \hat{\mathbf{s}}_t^m - \mathbf{s}_t^m \|_2^2 + \| \hat{\mathbf{p}}_t^m - \mathbf{p}_t^m \|_2^2 \right).
\end{align}
\vspace{-0.3cm}

\textit{Structural Masking ($\mathcal{L}_{struct}$).}
To simulate missing modalities, \rev{we randomly drop 1-3 entire modalities \qedit{at the sample level.}}
\qedit{The frozen tokenizer first provides target tokens for all modalities, after which the selected missing-modality tokens are excluded from the backbone input.
To prevent the model from over-focusing on absent sensors, we also apply random patch masking to the remaining observed modalities.}
\rev{When a single modality is missing ($|\mathcal{M}_{miss}|=1$),} we 
\qedit{reconstruct only its} modal-shared component ($\mathbf{s}^m$), while explicitly excluding \qedit{the non-identifiable} modal-specific component ($\mathbf{p}^m$). 
\rev{However, when 2 or 3 modalities are missing, we skip their reconstruction entirely (neither $\mathbf{s}^m$ nor $\mathbf{p}^m$), as even their shared components may be uninferable from the 
\qedit{limited observed context.}}
For the \qedit{observed} modalities ($m \in \mathcal{M}_{obs}$), the standard random masking loss applies. The structural objective is formulated as:

\vspace{-0.7cm}
\begin{multline}
\resizebox{0.9\linewidth}{!}{%
$\displaystyle \rev{\mathcal{L}_{struct} = \mathbb{I}(|\mathcal{M}_{miss}| = 1)}\cdot\left(\sum_{m \in \mathcal{M}_{miss}} \sum_{t=1}^{T_m} \|\hat{\mathbf{s}}_t^m - \mathbf{s}_t^m\|_2^2\right) + \sum_{m \in \mathcal{M}_{obs}} \sum_{t=1}^{T_m} \mathbb{M}_t^m \cdot \left( \|\hat{\mathbf{s}}_t^m - \mathbf{s}_t^m\|_2^2 + \|\hat{\mathbf{p}}_t^m - \mathbf{p}_t^m\|_2^2 \right),$%
}
\end{multline}

\rev{where $\mathbb{I(\cdot)}$is the indicator function.} \jw{Note that although the $\mathbf{s}$ of missing modality can be inferred from other modalities, we still reconstruct $\mathbf{s}$ in our objective. This encourages spatial consistency and enables the model to learn to handle missing modalities effectively.}

\textbf{Total Backbone Loss.}
The final optimization objective is the weighted sum of these two components:
\begin{align}
    \mathcal{L}_{backbone} = \mathcal{L}_{rand} + \lambda_{struct} \mathcal{L}_{struct},
\end{align}
where $\hat{s}$ and $\hat{p}$ denote the reconstructed embeddings predicted by the backbone, and $s, p$ are the ground-truth target embeddings frozen from the Orthogonal Tokenizer.
\vspace{-0.1cm}
\section{Experiments}

\subsection{Experiment Setup}

\textbf{Model Configuration.} 
\name uses a two-stage architecture \rev{(detailed in Appendix~\ref{appendix:hyperparameter})} with 50.7M parameters at inference: 16.94M in the orthogonal tokenizer and 33.74M in the missing-aware MoE backbone. 
All datasets are collected using the widely deployed Empatica E4 wristband, which records PPG at 64 Hz, EDA at 4 Hz, ACC at 32 Hz (3 axes), and TEMP at 4 Hz. We segment each stream into non-overlapping 10\,s windows and use a 1\,s patch size. 

\textbf{Datasets and Tasks.}
We use publicly available datasets for pretraining and downstream evaluation, with no subject or dataset overlap between the two stages. Pretraining uses three daily-life datasets: BIG IDEAs~\cite{bent2021engineering}, UE4W~\cite{hinkle2023fusion}, and the non-cannabis-user subset of CAN-STRESS~\cite{azghan2025can}, totaling 3{,}565 hours from 56 subjects. 
Downstream evaluation includes WESAD~\cite{schmidt2018introducing} for 3-class emotion classification (15 subjects), DaLiA~\cite{reiss2019deep} for 8-class activity classification (15 subjects), AAUWSS~\cite{djanian2025aauwss} for 5-class sleep-stage classification (13 subjects), \rev{and WEEE~\cite{gashi2022multidevice} for $\text{VO}_2$ prediction (17 subjects).}
\rev{We adopted a subject-wise 5-fold cross-validation.}
\rev{Specifically, neither the baselines nor \name have been exposed to the modality missingness in the downstream benchmark. Supervised training, fine-tuning, and linear probing are all conducted under the full-modal setting. During evaluation for a specific missing modality pattern, the designated modalities are removed from all test samples.} 

\textbf{Baselines and Metrics.} We compare \name with 3 supervised baselines and \rev{7} pretrained baselines, each configured with $\sim$50M parameters for capacity-matched comparison. The supervised baselines include ResNet~\cite{he2016resnet}, ViT~\cite{dosovitskiy2021vit}, and FlexMoE~\cite{yun2024flexmoe}. The pretrained baselines include SimCLR~\cite{chen2020simclr}, MSN~\cite{assran2022msn}, CMC~\cite{tian2020contrastive}, \rev{MF-CLR~\cite{duan2024mf}, and FOCAL~\cite{liu2023focal},} which are based on contrastive learning, as well as LSM~\cite{narayanswamy2025scaling} and LSM-2~\cite{xu2025lsm2}, which are masking-based methods specifically designed for wearable signals. Among them, FlexMoE and LSM-2 are designed to handle modality-missing scenarios.
\rev{For evaluation, we employ Macro-F1, Balanced Accuracy (BAcc), and Cohen's Kappa ($\kappa$) for classification, and Mean Absolute Error (MAE) alongside $R^2$ for regression.}
\qs{We report results under two standard transfer settings. \textbf{FT} denotes supervised end-to-end fine-tuning of the pretrained model on the downstream task. \textbf{LP} denotes linear probing, where the pretrained encoder is frozen, and only a linear classifier/regressor is trained on top.}








\begin{table*}[t] 
\centering
\vspace{-2ex}
\scriptsize 
\setlength{\tabcolsep}{4pt} 
\renewcommand{\arraystretch}{0.5}
\caption{\textbf{Full Modality Results.} 
\label{table:full}
Best results are marked in \textbf{bold}.
Second best results are \underline{underlined}.}
\resizebox{\linewidth}{!}{%
\begin{tabular}{lccccccccccc} 
\toprule
 & \multicolumn{3}{c}{WESAD} & \multicolumn{3}{c}{AAUWSS} & \multicolumn{3}{c}{DaLiA} & \multicolumn{2}{c}{WEEE} \\

\cmidrule(lr){2-4} \cmidrule(lr){5-7} \cmidrule(lr){8-10} \cmidrule(lr){11-12}
Method & $\uparrow$F$_1$ & $\uparrow$BAcc & $\uparrow\kappa$ & $\uparrow$F$_1$ & $\uparrow$BAcc & $\uparrow\kappa$ & $\uparrow$F$_1$ & $\uparrow$BAcc & $\uparrow\kappa$ & $\downarrow$MAE & $\uparrow$R$^2$ \\ 
\midrule

\multicolumn{12}{l}{\textit{\textbf{Supervised}}} \\
ResNet & \underline{.610}{\tiny$\pm.069$} & .635{\tiny$\pm.058$} & .561{\tiny$\pm.054$} & .236{\tiny$\pm.021$} & .254{\tiny$\pm.019$} & .110{\tiny$\pm.044$} & \underline{.733}{\tiny$\pm.004$} & \underline{.723}{\tiny$\pm.006$} & \underline{.656}{\tiny$\pm.008$} & 3.529{\tiny$\pm.123$} & .384{\tiny$\pm.049$} \\
ViT & .600{\tiny$\pm.041$} & \underline{.636}{\tiny$\pm.020$} & .588{\tiny$\pm.029$} & \underline{.251}{\tiny$\pm.095$} & \underline{.268}{\tiny$\pm.088$} & \underline{.169}{\tiny$\pm.170$} & .711{\tiny$\pm.016$} & .708{\tiny$\pm.012$} & .643{\tiny$\pm.010$} & 3.677{\tiny$\pm.154$} & .307{\tiny$\pm.062$} \\
FlexMoE & .555{\tiny$\pm.030$} & .612{\tiny$\pm.038$} & \underline{.609}{\tiny$\pm.070$} & .235{\tiny$\pm.022$} & .246{\tiny$\pm.030$} & .098{\tiny$\pm.050$} & .711{\tiny$\pm.006$} & .697{\tiny$\pm.008$} & .632{\tiny$\pm.007$} & \underline{3.039}{\tiny$\pm.203$} & \underline{.470}{\tiny$\pm.073$} \\
\rowcolor{grayrow} \textbf{Ours (FT)} & \textbf{.707}{\tiny$\pm.079$} & \textbf{.705}{\tiny$\pm.078$} & \textbf{.630}{\tiny$\pm.100$} & \textbf{.289}{\tiny$\pm.009$} & \textbf{.304}{\tiny$\pm.021$} & \textbf{.252}{\tiny$\pm.020$} & \textbf{.757}{\tiny$\pm.011$} & \textbf{.753}{\tiny$\pm.008$} & \textbf{.681}{\tiny$\pm.017$} & \textbf{2.633}{\tiny$\pm.109$} & \textbf{.668}{\tiny$\pm.030$} \\

\midrule 

\multicolumn{12}{l}{\textit{\textbf{Pretrained}}} \\
LSM     & .580{\tiny$\pm.001$} & \textbf{.651}{\tiny$\pm.002$} & \textbf{.666}{\tiny$\pm.005$} & .222{\tiny$\pm.022$} & .231{\tiny$\pm.015$} & .129{\tiny$\pm.058$} & .728{\tiny$\pm.004$} & .710{\tiny$\pm.005$} & .645{\tiny$\pm.007$} & 3.943{\tiny$\pm.079$} & .287{\tiny$\pm.026$} \\
LSM-2    & .533{\tiny$\pm.008$} & .559{\tiny$\pm.010$} & .515{\tiny$\pm.018$} & \underline{.281}{\tiny$\pm.007$} & \textbf{.316}{\tiny$\pm.005$} & \textbf{.339}{\tiny$\pm.007$} & .714{\tiny$\pm.002$} & .699{\tiny$\pm.002$} & .633{\tiny$\pm.005$} & \underline{3.711}{\tiny$\pm.010$} & \underline{.411}{\tiny$\pm.003$} \\
SimCLR  & .510{\tiny$\pm.017$} & .514{\tiny$\pm.028$} & .354{\tiny$\pm.017$} & .230{\tiny$\pm.005$} & .238{\tiny$\pm.003$} & .113{\tiny$\pm.028$} & .621{\tiny$\pm.013$} & .620{\tiny$\pm.023$} & .504{\tiny$\pm.014$} & 8.572{\tiny$\pm1.202$} & -3.796{\tiny$\pm2.688$} \\
MSN     & .512{\tiny$\pm.003$} & .569{\tiny$\pm.003$} & .522{\tiny$\pm.006$} & .206{\tiny$\pm.003$} & .225{\tiny$\pm.006$} & .179{\tiny$\pm.025$} & .728{\tiny$\pm.003$} & .711{\tiny$\pm.003$} & .649{\tiny$\pm.003$} & 3.970{\tiny$\pm.004$} & .297{\tiny$\pm.002$} \\
CMC     & .563{\tiny$\pm.008$} & .595{\tiny$\pm.014$} & .534{\tiny$\pm.026$} & .258{\tiny$\pm.005$} & .266{\tiny$\pm.005$} & .213{\tiny$\pm.024$} & .686{\tiny$\pm.005$} & .678{\tiny$\pm.007$} & .618{\tiny$\pm.006$} & 3.925{\tiny$\pm.295$} & .285{\tiny$\pm.076$} \\
MF-CLR   & .506{\tiny$\pm.001$} & .470{\tiny$\pm.002$} & .297{\tiny$\pm.003$} & .227{\tiny$\pm.004$} & .221{\tiny$\pm.001$} & .069{\tiny$\pm.002$} & .406{\tiny$\pm.002$} & .339{\tiny$\pm.001$} & .297{\tiny$\pm.001$} & 4.324{\tiny$\pm.038$} & .148{\tiny$\pm.018$} \\
FOCAL   & \underline{.584}{\tiny$\pm.004$} & .579{\tiny$\pm.011$} & .480{\tiny$\pm.025$} & .253{\tiny$\pm.003$} & .256{\tiny$\pm.003$} & .122{\tiny$\pm.008$} & \underline{.750}{\tiny$\pm.004$} & \underline{.741}{\tiny$\pm.004$} & \textbf{.688}{\tiny$\pm.005$} & 3.769{\tiny$\pm.213$} & .370{\tiny$\pm.076$} \\
\rowcolor{grayrow} \textbf{Ours (LP)} & \textbf{.640}{\tiny$\pm.042$} & \underline{.639}{\tiny$\pm.033$} & \underline{.590}{\tiny$\pm.037$} & \textbf{.287}{\tiny$\pm.006$} & \underline{.290}{\tiny$\pm.007$} & \underline{.248}{\tiny$\pm.025$} & \textbf{.752}{\tiny$\pm.003$} & \textbf{.746}{\tiny$\pm.004$} & \underline{.673}{\tiny$\pm.004$} & \textbf{2.606}{\tiny$\pm.067$} & \textbf{.693}{\tiny$\pm.018$} \\
\bottomrule
\end{tabular}}
\vspace{-0.6cm}
\end{table*}

\vspace{-0.2cm}
\subsection{Experimental Results}


\textbf{Results on Full Modalities.}
\rev{Table~\ref{table:full} compares} \name with existing baselines when all modalities are available.
\name consistently outperforms both supervised and pretrained baselines. For classification, \name improves macro-F1 by at least 0.097 on WESAD, 0.038 on AAUWSS, and 0.024 on DaLiA over the best supervised baselines, and by at least 0.060, 0.006, and \rev{0.002} on the respective datasets over the best pretrained baselines. \qedit{For regression}, \name reduces MAE by at least 0.406 and \rev{1.105}, while improving $\text{R}^2$ by 0.198 and \rev{0.282} over the best supervised and pretrained baselines on WEEE, respectively.
\qs{These results indicate that \name learns transferable representations across diverse health monitoring tasks. }
\qedit{Importantly, this full-modality performance indicates that robustness to missingness does not come at the cost of representation quality when all sensors are available; instead, preserving both modality-shared semantics and modality-specific residuals benefits downstream performance even when all sensors are available.}

\textbf{Results under \rev{Single} Modality Missing.} 
Table~\ref{table:missing} presents the model performance \qedit{when each modality is individually missing at test time across the four downstream tasks.}
\rev{Among the 32 primary metric comparisons (i.e., Macro-F1 for classification and MAE for regression $\times$ 4 missing-modality scenarios $\times$ 2 baseline types), \name outperforms the baselines in 23 cases. Across all 88 total reported metrics, \name achieves the best score in 65 cases and the second-best in 20 cases.}
\qs{We highlight a representative challenging setting where a critical sensor is missing. On DaLiA, removing ACC substantially degrades activity recognition for competing methods, whereas \name maintains F1 scores of 0.416 (FT) and 0.428 (LP). 
\rev{Moreover, when ACC is missing, \name’s margin over supervised baselines increases from 0.024 to 0.060, while its margin over pretrained baselines rises from 0.002 to 0.048,}
indicating that \name degrades more gracefully when a key sensor is unavailable.}

\begin{table*}[t!] 
\centering
\scriptsize
\setlength{\tabcolsep}{3pt} 
\renewcommand{\arraystretch}{0.5}
\caption{\textbf{Robustness under Single Modality Missing.} Comparison of performance when PPG, EDA, ACC, or TEMP is missing.
Best results are marked in \textbf{bold}, second best are \underline{underlined}.}
\label{table:missing}

\resizebox{\linewidth}{!}{%
\begin{tabular}{lccccccccccc}
\toprule
 & \multicolumn{3}{c}{WESAD} & \multicolumn{3}{c}{AAUWSS} & \multicolumn{3}{c}{DaLiA} & \multicolumn{2}{c}{WEEE} \\
\cmidrule(lr){2-4} \cmidrule(lr){5-7} \cmidrule(lr){8-10} \cmidrule(lr){11-12}
Method & $\uparrow$F$_1$ & $\uparrow$BAcc & $\uparrow\kappa$ & $\uparrow$F$_1$ & $\uparrow$BAcc & $\uparrow\kappa$ & $\uparrow$F$_1$ & $\uparrow$BAcc & $\uparrow\kappa$ & $\downarrow$MAE & $\uparrow$R$^2$ \\ 
\midrule

\multicolumn{12}{c}{\cellcolor{gray!20}\textbf{PPG Missing}} \\
\midrule
\multicolumn{12}{l}{\textit{Supervised}} \\
ResNet & \underline{.618}{\tiny$\pm.064$} & \underline{.643}{\tiny$\pm.053$} & .574{\tiny$\pm.050$} & .219{\tiny$\pm.021$} & .259{\tiny$\pm.017$} & .099{\tiny$\pm.035$} & \textbf{.735}{\tiny$\pm.007$} & \underline{.728}{\tiny$\pm.007$} & \underline{.661}{\tiny$\pm.012$} & \underline{3.405}{\tiny$\pm.132$} & \underline{.415}{\tiny$\pm.038$} \\
ViT & .590{\tiny$\pm.037$} & .627{\tiny$\pm.018$} & .578{\tiny$\pm.027$} & \underline{.247}{\tiny$\pm.067$} & \underline{.271}{\tiny$\pm.064$} & \underline{.153}{\tiny$\pm.148$} & .686{\tiny$\pm.022$} & .685{\tiny$\pm.014$} & .629{\tiny$\pm.023$} & 3.630{\tiny$\pm.103$} & .306{\tiny$\pm.055$} \\
FlexMoE & .562{\tiny$\pm.029$} & .621{\tiny$\pm.036$} & \underline{.625}{\tiny$\pm.067$} & .192{\tiny$\pm.029$} & .229{\tiny$\pm.027$} & .052{\tiny$\pm.026$} & .717{\tiny$\pm.010$} & .705{\tiny$\pm.015$} & .641{\tiny$\pm.011$} & 3.837{\tiny$\pm.342$} & .222{\tiny$\pm.135$} \\
\rowcolor{grayrow} \textbf{Ours (FT)} & \textbf{.727}{\tiny$\pm.030$} & \textbf{.729}{\tiny$\pm.045$} & \textbf{.631}{\tiny$\pm.047$} & \textbf{.250}{\tiny$\pm.006$} & \textbf{.277}{\tiny$\pm.006$} & \textbf{.161}{\tiny$\pm.008$} & \underline{.716}{\tiny$\pm.048$} & \textbf{.741}{\tiny$\pm.027$} & \textbf{.663}{\tiny$\pm.028$} & \textbf{3.152}{\tiny$\pm.232$} & \textbf{.541}{\tiny$\pm.061$} \\
\multicolumn{12}{l}{\textit{Pretrained}} \\
LSM     & .369{\tiny$\pm.032$} & .418{\tiny$\pm.026$} & .211{\tiny$\pm.061$} & .181{\tiny$\pm.020$} & .208{\tiny$\pm.014$} & .028{\tiny$\pm.046$} & .622{\tiny$\pm.017$} & .645{\tiny$\pm.013$} & .540{\tiny$\pm.014$} & 4.062{\tiny$\pm.165$} & .272{\tiny$\pm.047$} \\
LSM-2    & \underline{.510}{\tiny$\pm.002$} & .541{\tiny$\pm.003$} & .473{\tiny$\pm.005$} & .226{\tiny$\pm.008$} & \textbf{.296}{\tiny$\pm.020$} & \textbf{.203}{\tiny$\pm.026$} & .688{\tiny$\pm.022$} & .669{\tiny$\pm.024$} & .599{\tiny$\pm.023$} & 4.514{\tiny$\pm.026$} & .122{\tiny$\pm.009$} \\
SimCLR  & .507{\tiny$\pm.019$} & .511{\tiny$\pm.031$} & .349{\tiny$\pm.018$} & \underline{.230}{\tiny$\pm.005$} & .237{\tiny$\pm.003$} & .114{\tiny$\pm.030$} & .621{\tiny$\pm.012$} & .620{\tiny$\pm.022$} & .506{\tiny$\pm.010$} & \underline{3.685}{\tiny$\pm.132$} & \underline{.338}{\tiny$\pm.059$} \\
MSN     & .495{\tiny$\pm.002$} & \underline{.548}{\tiny$\pm.002$} & \underline{.484}{\tiny$\pm.004$} & .189{\tiny$\pm.001$} & .204{\tiny$\pm.001$} & .076{\tiny$\pm.006$} & \underline{.704}{\tiny$\pm.023$} & \underline{.674}{\tiny$\pm.022$} & \underline{.630}{\tiny$\pm.023$} & 4.008{\tiny$\pm.006$} & .297{\tiny$\pm.002$} \\
CMC     & .501{\tiny$\pm.053$} & .541{\tiny$\pm.044$} & .409{\tiny$\pm.098$} & .190{\tiny$\pm.007$} & .210{\tiny$\pm.006$} & .074{\tiny$\pm.032$} & .283{\tiny$\pm.066$} & .357{\tiny$\pm.065$} & .248{\tiny$\pm.064$} & 5.035{\tiny$\pm.802$} & -.114{\tiny$\pm.257$} \\
MF-CLR   & .489{\tiny$\pm.002$} & .513{\tiny$\pm.014$} & .391{\tiny$\pm.027$} & .195{\tiny$\pm.011$} & .245{\tiny$\pm.002$} & \underline{.170}{\tiny$\pm.012$} & .291{\tiny$\pm.001$} & .313{\tiny$\pm.001$} & .270{\tiny$\pm.001$} & 4.281{\tiny$\pm.124$} & .250{\tiny$\pm.039$} \\
FOCAL   & .347{\tiny$\pm.131$} & .402{\tiny$\pm.084$} & .130{\tiny$\pm.150$} & .176{\tiny$\pm.036$} & .222{\tiny$\pm.017$} & .052{\tiny$\pm.037$} & .109{\tiny$\pm.025$} & .162{\tiny$\pm.019$} & .033{\tiny$\pm.017$} & 26.241{\tiny$\pm1.714$} & -19.995{\tiny$\pm2.733$} \\
\rowcolor{grayrow} \textbf{Ours (LP)} & \textbf{.626}{\tiny$\pm.042$} & \textbf{.635}{\tiny$\pm.027$} & \textbf{.585}{\tiny$\pm.024$} & \textbf{.251}{\tiny$\pm.006$} & \underline{.281}{\tiny$\pm.008$} & .102{\tiny$\pm.008$} & \textbf{.706}{\tiny$\pm.039$} & \textbf{.721}{\tiny$\pm.031$} & \textbf{.639}{\tiny$\pm.028$} & \textbf{3.191}{\tiny$\pm.056$} & \textbf{.497}{\tiny$\pm.029$} \\

\midrule
\multicolumn{12}{c}{\cellcolor{gray!20}\textbf{EDA Missing}} \\
\midrule
\multicolumn{12}{l}{\textit{Supervised}} \\
ResNet & .512{\tiny$\pm.038$} & .537{\tiny$\pm.031$} & \underline{.393}{\tiny$\pm.036$} & .208{\tiny$\pm.013$} & .229{\tiny$\pm.016$} & .057{\tiny$\pm.026$} & .486{\tiny$\pm.020$} & .491{\tiny$\pm.014$} & .376{\tiny$\pm.022$} & 6.223{\tiny$\pm.653$} & -.726{\tiny$\pm.334$} \\
ViT & \underline{.554}{\tiny$\pm.041$} & \underline{.593}{\tiny$\pm.037$} & \textbf{.475}{\tiny$\pm.055$} & .217{\tiny$\pm.090$} & .233{\tiny$\pm.072$} & .060{\tiny$\pm.152$} & .502{\tiny$\pm.024$} & .500{\tiny$\pm.016$} & .412{\tiny$\pm.012$} & 5.848{\tiny$\pm.901$} & -.530{\tiny$\pm.451$} \\
FlexMoE & .356{\tiny$\pm.055$} & .398{\tiny$\pm.049$} & .156{\tiny$\pm.108$} & \textbf{.253}{\tiny$\pm.029$} & \underline{.261}{\tiny$\pm.027$} & \underline{.089}{\tiny$\pm.038$} & \underline{.553}{\tiny$\pm.025$} & \underline{.576}{\tiny$\pm.027$} & \underline{.475}{\tiny$\pm.038$} & \underline{3.171}{\tiny$\pm.234$} & \underline{.458}{\tiny$\pm.075$} \\
\rowcolor{grayrow} \textbf{Ours (FT)} & \textbf{.567}{\tiny$\pm.048$} & \underline{.588}{\tiny$\pm.038$} & .375{\tiny$\pm.317$} & \underline{.249}{\tiny$\pm.009$} & \textbf{.278}{\tiny$\pm.010$} & \textbf{.110}{\tiny$\pm.014$} & \textbf{.711}{\tiny$\pm.040$} & \textbf{.684}{\tiny$\pm.049$} & \textbf{.581}{\tiny$\pm.044$}& \textbf{2.985}{\tiny$\pm.132$} & \textbf{.578}{\tiny$\pm.024$} \\
\multicolumn{12}{l}{\textit{Pretrained}} \\
LSM     & .506{\tiny$\pm.014$} & .556{\tiny$\pm.019$} & .484{\tiny$\pm.034$} & .187{\tiny$\pm.017$} & .210{\tiny$\pm.009$} & .047{\tiny$\pm.047$} & .456{\tiny$\pm.026$} & .491{\tiny$\pm.018$} & .395{\tiny$\pm.020$} & 6.291{\tiny$\pm.088$} & -.457{\tiny$\pm.046$} \\
LSM-2    & .253{\tiny$\pm.008$} & .337{\tiny$\pm.005$} & .021{\tiny$\pm.013$} & .217{\tiny$\pm.008$} & .250{\tiny$\pm.003$} & \textbf{.192}{\tiny$\pm.027$} & .514{\tiny$\pm.018$} & .543{\tiny$\pm.011$} & .440{\tiny$\pm.022$} & \underline{5.422}{\tiny$\pm.027$} & \underline{-.116}{\tiny$\pm.007$} \\
SimCLR  & .445{\tiny$\pm.025$} & .451{\tiny$\pm.026$} & .240{\tiny$\pm.038$} & .175{\tiny$\pm.022$} & .204{\tiny$\pm.017$} & .049{\tiny$\pm.056$} & .461{\tiny$\pm.036$} & .463{\tiny$\pm.046$} & .331{\tiny$\pm.043$} & 6.201{\tiny$\pm.316$} & -.489{\tiny$\pm.163$} \\
MSN     & \underline{.515}{\tiny$\pm.007$} & \underline{.573}{\tiny$\pm.010$} & \textbf{.520}{\tiny$\pm.018$} & .197{\tiny$\pm.003$} & .212{\tiny$\pm.004$} & .119{\tiny$\pm.016$} & .481{\tiny$\pm.010$} & .512{\tiny$\pm.009$} & .404{\tiny$\pm.010$} & 6.553{\tiny$\pm.027$} & -.633{\tiny$\pm.014$} \\
CMC     & .465{\tiny$\pm.062$} & .517{\tiny$\pm.065$} & .397{\tiny$\pm.134$} & .213{\tiny$\pm.008$} & .236{\tiny$\pm.017$} & \underline{.152}{\tiny$\pm.016$} & .352{\tiny$\pm.091$} & .426{\tiny$\pm.094$} & .269{\tiny$\pm.085$} & 10.225{\tiny$\pm4.296$} & -3.229{\tiny$\pm3.313$} \\
MF-CLR   & .345{\tiny$\pm.004$} & .336{\tiny$\pm.007$} & .005{\tiny$\pm.017$} & .177{\tiny$\pm.001$} & .199{\tiny$\pm.000$} & -.003{\tiny$\pm.001$} & .151{\tiny$\pm.004$} & .139{\tiny$\pm.000$} & .018{\tiny$\pm.001$} & 5.446{\tiny$\pm.069$} & -.193{\tiny$\pm.038$} \\
FOCAL   & .436{\tiny$\pm.022$} & .442{\tiny$\pm.029$} & .231{\tiny$\pm.042$} & \textbf{.254}{\tiny$\pm.005$} & \underline{.261}{\tiny$\pm.013$} & .095{\tiny$\pm.005$} & \underline{.677}{\tiny$\pm.028$} & \underline{.666}{\tiny$\pm.027$} & \underline{.575}{\tiny$\pm.027$} & 5.550{\tiny$\pm1.671$} & -.336{\tiny$\pm.748$} \\
\rowcolor{grayrow} \textbf{Ours (LP)} & \textbf{.585}{\tiny$\pm.053$} & \textbf{.600}{\tiny$\pm.046$} & \underline{.494}{\tiny$\pm.053$} & \underline{.241}{\tiny$\pm.009$} & \textbf{.268}{\tiny$\pm.010$} & .073{\tiny$\pm.014$} & \textbf{.721}{\tiny$\pm.027$} & \textbf{.701}{\tiny$\pm.027$} & \textbf{.600}{\tiny$\pm.030$} & \textbf{2.925}{\tiny$\pm.049$} & \textbf{.615}{\tiny$\pm.020$} \\

\midrule
\multicolumn{12}{c}{\cellcolor{gray!20}\textbf{ACC Missing}} \\
\midrule
\multicolumn{12}{l}{\textit{Supervised}} \\
ResNet & .460{\tiny$\pm.033$} & .500{\tiny$\pm.035$} & .250{\tiny$\pm.094$} & .176{\tiny$\pm.049$} & .186{\tiny$\pm.058$} & -.009{\tiny$\pm.154$} & .255{\tiny$\pm.033$} & .354{\tiny$\pm.029$} & .237{\tiny$\pm.051$} & 4.410{\tiny$\pm.497$} & -.033{\tiny$\pm.202$} \\
ViT & .472{\tiny$\pm.043$} & .523{\tiny$\pm.057$} & .388{\tiny$\pm.118$} & .200{\tiny$\pm.037$} & .209{\tiny$\pm.044$} & .050{\tiny$\pm.067$} & .268{\tiny$\pm.041$} & .346{\tiny$\pm.040$} & .200{\tiny$\pm.036$} & 4.428{\tiny$\pm1.097$} & .000{\tiny$\pm.415$} \\
FlexMoE & \underline{.505}{\tiny$\pm.012$} & \underline{.556}{\tiny$\pm.020$} & \underline{.503}{\tiny$\pm.026$} & \underline{.216}{\tiny$\pm.023$} & \underline{.233}{\tiny$\pm.015$} & \underline{.114}{\tiny$\pm.046$} & \underline{.356}{\tiny$\pm.052$} & \underline{.427}{\tiny$\pm.025$} & \underline{.352}{\tiny$\pm.084$} & \underline{3.045}{\tiny$\pm.309$} & \underline{.497}{\tiny$\pm.093$} \\
\rowcolor{grayrow} \textbf{Ours (FT)} & \textbf{.618}{\tiny$\pm.075$} & \textbf{.642}{\tiny$\pm.218$} & \textbf{.513}{\tiny$\pm.078$} & \textbf{.239}{\tiny$\pm.006$} & \textbf{.269}{\tiny$\pm.009$} & \textbf{.170}{\tiny$\pm.023$} & \textbf{.416}{\tiny$\pm.091$} & \textbf{.452}{\tiny$\pm.076$} & \textbf{.414}{\tiny$\pm.082$} & \textbf{2.797}{\tiny$\pm.094$} & \textbf{.603}{\tiny$\pm.026$} \\
\multicolumn{12}{l}{\textit{Pretrained}} \\
LSM     & .561{\tiny$\pm.005$} & \textbf{.636}{\tiny$\pm.004$} & \textbf{.626}{\tiny$\pm.009$} & .221{\tiny$\pm.017$} & .233{\tiny$\pm.015$} & .143{\tiny$\pm.057$} & .355{\tiny$\pm.045$} & .387{\tiny$\pm.032$} & .337{\tiny$\pm.039$} & 3.759{\tiny$\pm.160$} & .354{\tiny$\pm.058$} \\
LSM-2    & .510{\tiny$\pm.006$} & .556{\tiny$\pm.007$} & .502{\tiny$\pm.016$} & .213{\tiny$\pm.018$} & \textbf{.317}{\tiny$\pm.012$} & .238{\tiny$\pm.040$} & \underline{.380}{\tiny$\pm.014$} & \underline{.445}{\tiny$\pm.012$} & \textbf{.422}{\tiny$\pm.024$} & \underline{3.411}{\tiny$\pm.003$} & \underline{.479}{\tiny$\pm.001$} \\
SimCLR  & .245{\tiny$\pm.098$} & .350{\tiny$\pm.099$} & .026{\tiny$\pm.157$} & .133{\tiny$\pm.000$} & .200{\tiny$\pm.000$} & .000{\tiny$\pm.000$} & .038{\tiny$\pm.028$} & .139{\tiny$\pm.029$} & .019{\tiny$\pm.039$} & 19.524{\tiny$\pm13.451$} & -16.090{\tiny$\pm19.537$} \\
MSN     & .284{\tiny$\pm.022$} & .357{\tiny$\pm.011$} & .094{\tiny$\pm.042$} & .230{\tiny$\pm.003$} & .258{\tiny$\pm.008$} & \underline{.265}{\tiny$\pm.025$} & .290{\tiny$\pm.040$} & .332{\tiny$\pm.030$} & .291{\tiny$\pm.027$} & 6.270{\tiny$\pm.031$} & -1.036{\tiny$\pm.024$} \\
CMC     & .483{\tiny$\pm.032$} & .519{\tiny$\pm.032$} & .370{\tiny$\pm.060$} & \underline{.239}{\tiny$\pm.007$} & .254{\tiny$\pm.004$} & .198{\tiny$\pm.013$} & .317{\tiny$\pm.029$} & .387{\tiny$\pm.027$} & .342{\tiny$\pm.046$} & 5.729{\tiny$\pm.679$} & -.521{\tiny$\pm.355$} \\
MF-CLR   & .502{\tiny$\pm.004$} & .477{\tiny$\pm.002$} & .315{\tiny$\pm.006$} & .225{\tiny$\pm.002$} & .218{\tiny$\pm.001$} & .067{\tiny$\pm.001$} & .377{\tiny$\pm.005$} & .294{\tiny$\pm.001$} & .250{\tiny$\pm.002$} & 4.329{\tiny$\pm.058$} & .147{\tiny$\pm.020$} \\
FOCAL   & \underline{.598}{\tiny$\pm.069$} & \underline{.626}{\tiny$\pm.036$} & .447{\tiny$\pm.099$} & .211{\tiny$\pm.001$} & .226{\tiny$\pm.004$} & .125{\tiny$\pm.007$} & .249{\tiny$\pm.032$} & .321{\tiny$\pm.034$} & .240{\tiny$\pm.056$} & 4.516{\tiny$\pm.769$} & .111{\tiny$\pm.276$} \\
\rowcolor{grayrow} \textbf{Ours (LP)} & \textbf{.610}{\tiny$\pm.030$} & .608{\tiny$\pm.023$} & \underline{.514}{\tiny$\pm.028$} & \textbf{.272}{\tiny$\pm.006$} & \underline{.270}{\tiny$\pm.009$} & \textbf{.300}{\tiny$\pm.023$} & \textbf{.428}{\tiny$\pm.017$} & \textbf{.458}{\tiny$\pm.014$} & \underline{.405}{\tiny$\pm.017$} & \textbf{3.358}{\tiny$\pm.041$} & \textbf{.545}{\tiny$\pm.017$} \\

\midrule
\multicolumn{12}{c}{\cellcolor{gray!20}\textbf{TEMP Missing}} \\
\midrule
\multicolumn{12}{l}{\textit{Supervised}} \\
ResNet & .490{\tiny$\pm.057$} & .509{\tiny$\pm.056$} & .338{\tiny$\pm.065$} & .198{\tiny$\pm.016$} & .220{\tiny$\pm.021$} & .033{\tiny$\pm.029$} & \underline{.571}{\tiny$\pm.035$} & \underline{.567}{\tiny$\pm.025$} & \underline{.489}{\tiny$\pm.040$} & 3.792{\tiny$\pm.225$} & .338{\tiny$\pm.059$} \\
ViT & \underline{.553}{\tiny$\pm.058$} & .580{\tiny$\pm.044$} & .486{\tiny$\pm.063$} & \textbf{.235}{\tiny$\pm.049$} & \underline{.256}{\tiny$\pm.023$} & \underline{.130}{\tiny$\pm.107$} & .504{\tiny$\pm.026$} & .510{\tiny$\pm.024$} & .420{\tiny$\pm.020$} & 3.913{\tiny$\pm.242$} & .278{\tiny$\pm.107$} \\
FlexMoE & .531{\tiny$\pm.021$} & \underline{.592}{\tiny$\pm.027$} & .535{\tiny$\pm.039$} & .208{\tiny$\pm.019$} & \underline{.256}{\tiny$\pm.030$} & .099{\tiny$\pm.055$} & .516{\tiny$\pm.050$} & .518{\tiny$\pm.045$} & .450{\tiny$\pm.048$} & \underline{3.109}{\tiny$\pm.278$} & \underline{.478}{\tiny$\pm.086$} \\
\rowcolor{grayrow} \textbf{Ours (FT)} & \textbf{.566}{\tiny$\pm.158$} & \textbf{.608}{\tiny$\pm.024$} & \underline{.510}{\tiny$\pm.063$} & \underline{.234}{\tiny$\pm.034$} & \textbf{.270}{\tiny$\pm.023$} & \textbf{.132}{\tiny$\pm.059$} & \textbf{.648}{\tiny$\pm.014$} & \textbf{.640}{\tiny$\pm.017$} & \textbf{.553}{\tiny$\pm.013$} & \textbf{2.722}{\tiny$\pm.169$} & \textbf{.639}{\tiny$\pm.040$} \\
\multicolumn{12}{l}{\textit{Pretrained}} \\
LSM     & .536{\tiny$\pm.015$} & \textbf{.604}{\tiny$\pm.020$} & \textbf{.573}{\tiny$\pm.034$} & .210{\tiny$\pm.017$} & .222{\tiny$\pm.012$} & .088{\tiny$\pm.040$} & .566{\tiny$\pm.007$} & .550{\tiny$\pm.011$} & .477{\tiny$\pm.011$} & 3.869{\tiny$\pm.128$} & .297{\tiny$\pm.045$} \\
LSM-2    & \textbf{.556}{\tiny$\pm.019$} & \underline{.570}{\tiny$\pm.016$} & \underline{.514}{\tiny$\pm.023$} & .227{\tiny$\pm.009$} & .230{\tiny$\pm.010$} & .127{\tiny$\pm.025$} & .458{\tiny$\pm.108$} & .475{\tiny$\pm.079$} & .424{\tiny$\pm.077$} & 3.842{\tiny$\pm.012$} & .378{\tiny$\pm.003$} \\
SimCLR  & .440{\tiny$\pm.023$} & .444{\tiny$\pm.023$} & .251{\tiny$\pm.039$} & .161{\tiny$\pm.013$} & .167{\tiny$\pm.014$} & -.048{\tiny$\pm.037$} & .401{\tiny$\pm.047$} & .409{\tiny$\pm.050$} & .327{\tiny$\pm.042$} & 4.112{\tiny$\pm.135$} & .222{\tiny$\pm.059$} \\
MSN     & .518{\tiny$\pm.013$} & .556{\tiny$\pm.008$} & .487{\tiny$\pm.010$} & .197{\tiny$\pm.001$} & .236{\tiny$\pm.007$} & .128{\tiny$\pm.014$} & .389{\tiny$\pm.021$} & .397{\tiny$\pm.016$} & .342{\tiny$\pm.013$} & 4.028{\tiny$\pm.002$} & .343{\tiny$\pm.001$} \\
CMC     & .515{\tiny$\pm.015$} & .544{\tiny$\pm.022$} & .414{\tiny$\pm.037$} & \underline{.237}{\tiny$\pm.011$} & \textbf{.281}{\tiny$\pm.010$} & \underline{.148}{\tiny$\pm.030$} & .591{\tiny$\pm.019$} & .597{\tiny$\pm.016$} & .514{\tiny$\pm.023$} & 3.862{\tiny$\pm.619$} & .273{\tiny$\pm.175$} \\
MF-CLR   & .492{\tiny$\pm.003$} & .470{\tiny$\pm.002$} & .297{\tiny$\pm.003$} & .226{\tiny$\pm.004$} & .221{\tiny$\pm.001$} & .069{\tiny$\pm.002$} & .291{\tiny$\pm.005$} & .339{\tiny$\pm.001$} & .297{\tiny$\pm.001$} & 4.752{\tiny$\pm.035$} & .009{\tiny$\pm.020$} \\
FOCAL   & .534{\tiny$\pm.027$} & .541{\tiny$\pm.033$} & .433{\tiny$\pm.056$} & .232{\tiny$\pm.003$} & .235{\tiny$\pm.005$} & .045{\tiny$\pm.009$} & \textbf{.660}{\tiny$\pm.007$} & \textbf{.653}{\tiny$\pm.005$} & \textbf{.558}{\tiny$\pm.014$} & \underline{3.527}{\tiny$\pm.465$} & \underline{.423}{\tiny$\pm.158$} \\
\rowcolor{grayrow} \textbf{Ours (LP)} & \underline{.540}{\tiny$\pm.022$} & .537{\tiny$\pm.012$} & .438{\tiny$\pm.023$} & \textbf{.258}{\tiny$\pm.004$} & \underline{.264}{\tiny$\pm.004$} & \textbf{.193}{\tiny$\pm.014$} & \underline{.625}{\tiny$\pm.026$} & \underline{.618}{\tiny$\pm.011$} & \underline{.540}{\tiny$\pm.015$} & \textbf{2.681}{\tiny$\pm.108$} & \textbf{.675}{\tiny$\pm.027$} \\

\bottomrule
\end{tabular}}
\vspace{-0.5cm}
\end{table*}
\textbf{\rev{Results under Multiple Modality Missing.}}
To evaluate robustness under severe sensor absence, we test all cases where 2 or 3 modalities are simultaneously missing, as detailed in Appendix~\ref{multiple_missing}.
\rev{Across 3 classification tasks \qedit{(10 missing patterns and 30 evaluation settings)}, \name ranks Top-1 in 20/30 and Top-2 in 25/30 against pretrained baselines, and Top-1 in 24/30 and Top-2 in 28/30 against supervised baselines. For regression, \name ranks Top-1 in 7/10 (MAE) and 7/10 settings ($\text{R}^2$) against pretrained baselines, and achieves Top-1 in 8/10 (MAE) and 7/10 ($\text{R}^2$) against supervised methods.} 
\qedit{These results show that \name remains robust across all multiple missing-modality settings.}



\begin{figure}[h]
    \centering 
    \vspace{-0.4cm}
    \setlength{\abovecaptionskip}{1pt}  
    \includegraphics[width=0.98\linewidth]{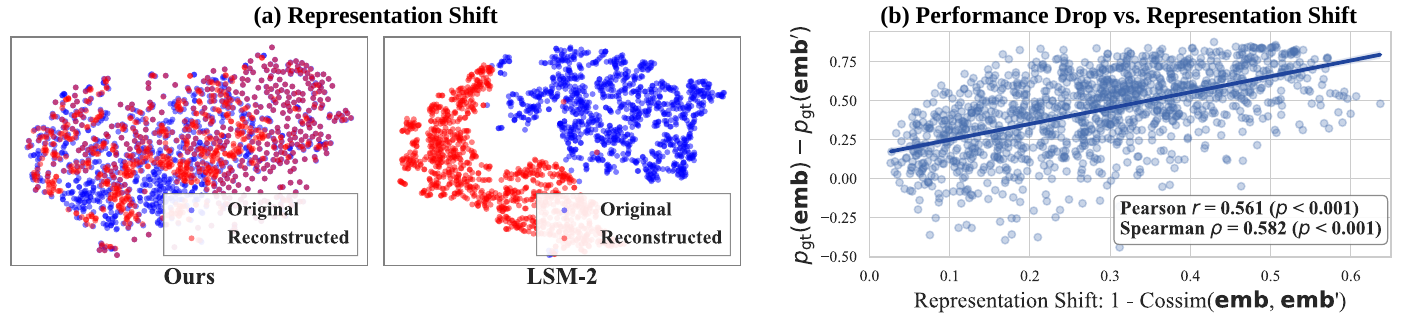}
    \vspace{-1ex}
    \caption{\textbf{(a)} t-SNE visualization of the embeddings of the original samples and those with reconstructed ACC signals. \textbf{(b)} \rev{Performance Drop vs. Representation Shift in LSM-2.}} 
    \label{fig:hallucination_vis} 
    \vspace{-3ex}
\end{figure}

\textbf{Case Study of Hallucination Effects.} \name remains robust when key modalities are missing, 
\rev{primarily because the Orthogonal Tokenizer enforces strict information boundaries, which enables the Missing-aware MoE Backbone to learn how to accurately recover missing modality information, rather than relying on uncertain guesses that may lead to hallucinations.}
\qedit{We support this intuition through a comparison with LSM-2}
, which directly reconstructs the \rev{entire} raw data including both modality-shared and \rev{uninferable} modality-specific components \rev{when crucial modality is missing, leading to semantic inconsistency.}
\qs{To visualize these effects, we use the DaLiA dataset. We freeze the encoders of \name and LSM-2 and train an auxiliary decoder (without masking) to map global embeddings back to raw signals.  For a full-modality sample $X=\{\mathbf{x}^{PPG}, \mathbf{x}^{EDA}, \mathbf{x}^{ACC}, \mathbf{x}^{TEMP}\}$, we obtain the reference embedding $\mathbf{emb}$ by feeding $X$ into the frozen encoder. We then drop ACC, run the full pipeline to reconstruct $\hat{\mathbf{x}}^{ACC}$, form $X'=\{\mathbf{x}^{PPG}, \mathbf{x}^{EDA}, \hat{\mathbf{x}}^{ACC}, \mathbf{x}^{TEMP}\}$, and compute $\mathbf{emb'}$ from $X'$ using the same encoder. If the reconstruction introduces unsupported modality-specific details, \qedit{the resulting representation should deviate from the original one beyond what is justified by the observed context.}
We therefore visualize the t-SNE distributions of $\mathbf{emb}$ and $\mathbf{emb'}$ to quantify the induced representation shift.}
As shown in Figure~\ref{fig:hallucination_vis}(a), LSM-2 produces 
a larger embedding shift when ACC is missing. The cosine similarity between the original and reconstructed embeddings is 0.732 for LSM-2, compared to 0.813 for \name. 
\rev{Furthermore, per-sample analysis in Figure~\ref{fig:hallucination_vis}(b) reveals a significant positive correlation (Pearson $r=0.561$, $p<0.001$) between representation shift and true class probability drop. These results support that reconstructing uninferable modality-specific details can induce hallucination-like completions, causing semantic inconsistency that directly degrades downstream performance.} 
\vspace{-0.2cm}
\subsection{Analysis}
\vspace{-0.2cm}
\textbf{Quantitative Assessment of Disentanglement.}
To quantify the independence between the modal-shared component $\mathbf{s}$ and the modal-specific component $\mathbf{p}$, we compute the Hilbert-Schmidt Independence Criterion (HSIC) using a Gaussian kernel.
Experimental results show that
the soft disentanglement baseline FOCAL~\cite{liu2023focal}, which relies on orthogonal loss, struggles to decouple low-frequency modalities such as EDA and TEMP. In contrast, \name consistently achieves HSIC values between 0.003 and 0.005 across all modalities, with an average less than 40\% of that of the baseline. 
Unlike typical methods that require explicit disentanglement losses
\begin{wraptable}{r}{0.35\textwidth} 
    \centering
    \vspace{-1ex} 
    \tiny
    \setlength{\tabcolsep}{2pt}
    \caption{Independence evaluation.}
    \label{tab:hsic_results}
    \begin{tabular}{lccccc}
        \toprule
        \multirow{2}{*}{\textbf{Method}} & \multicolumn{5}{c}{\textbf{HSIC (Gaussian Kernel) $\downarrow$}} \\
        \cmidrule(lr){2-6}
         & PPG & EDA & ACC & TEMP & \textbf{Average} \\
        \midrule
        FOCAL & 0.0045 & 0.0140 & 0.0076 & 0.0174 & 0.0108 \\
        Ours & \textbf{0.0039} & \textbf{0.0046} & \textbf{0.0033} & \textbf{0.0048} & \textbf{0.0042} \\
        \bottomrule
    \end{tabular}
    \vspace{-4ex} 
\end{wraptable}
between s and p, \name achieves effective disentanglement purely through the geometric design of the tokenizer. Moreover, 
\name demonstrates consistent robustness across different modalities, preventing high-frequency signals from dominating the disentanglement process, 
which is important for wearable data with large variations in sampling rate and spectral content.
    
\begin{wrapfigure}{r}{0.48\textwidth} 
    \centering
    \vspace{-2ex}
    \includegraphics[width=\linewidth, trim=0.2cm 0.2cm 0.cm 0cm, clip]{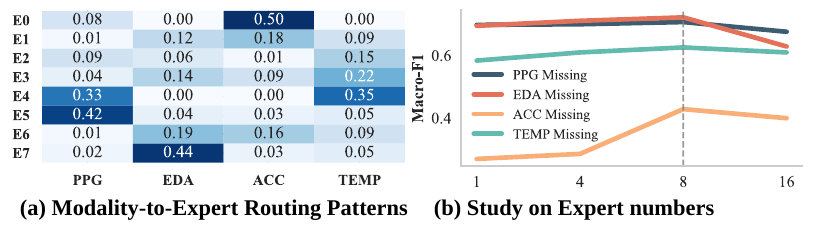}
    \captionsetup{skip=5pt}
    \caption{Analysis of the MoE Mechanism.} 
    \label{fig:moe_vis} 
    \vspace{-2ex}
\end{wrapfigure}

\textbf{Expert Specialization and Capacity Analysis.}
Figure~\ref{fig:moe_vis}(a) shows that the router \qedit{learns modality- and availability-dependent expert activation patterns.}
Taking EDA as an example,
Experts 1–3 capture different
dependencies between EDA and the other three modalities, while Expert 7 primarily handles cases where other modalities are missing. 
Figure~\ref{fig:moe_vis}(b) further analyzes the effect of the number of experts on DaLiA.
Across different modality missingness scenarios, using 8 experts yields the best performance. When the number of experts is set to 1, the 
\qedit{backbone degenerates into a}
dense architecture \qedit{and performance drops}, highlighting the necessity of the MoE structure for handling modality missingness effectively.

\begin{figure}[h]
    \centering 
    \vspace{-1ex}
    \setlength{\abovecaptionskip}{1pt}  
    \includegraphics[width=0.98\linewidth]{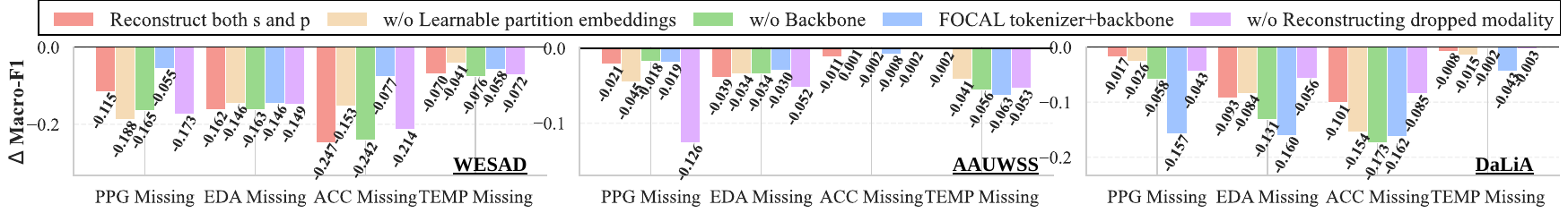}
    \caption{Macro-F1 drop of different ablations compared to \name. }
    \label{fig:ablation_objective} 
    \vspace{-0.5cm}
\end{figure}
\para{Ablation Studies.} We conduct three groups of ablations to assess the contribution of each design choice. 1) replacing the structural disentanglement in the orthogonal tokenizer with a FOCAL-type soft disentanglement based on orthogonal loss; 2) removing the missing-aware MoE backbone or replacing the learnable partition embedding with a fixed sinusoidal encoding; and 3) modifying the objective for structurally missing modalities from reconstructing only shared components to reconstructing both or reconstructing neither of the shared and modality-specific components.
\qs{Figure~\ref{fig:ablation_objective} reports the \textbf{Macro-F1 drop} of each ablated variant relative to the full \name pipeline. Each ablation causes a clear degradation across all three tasks and all four missing-modality settings (with only 1 exception among 60 metrics). Overall, the results suggest that soft disentanglement allows leakage between shared and modality-specific factors, weakening information separation; the missing-aware MoE backbone is necessary to adapt computation to heterogeneous modality availability, whereas fixed encodings are insufficient to capture modality-dependent semantic relations (e.g., the semantic affinity between EDA and TEMP may differ from that between EDA and ACC); and reconstructing modality-specific components for missing modalities incentivizes the model to generate unobserved content, increasing the risk of hallucination.}
Despite $\mathbf{s}$ being partially predictable from the observed modalities, explicitly reconstructing $\mathbf{s}$ improves robustness relative to skipping reconstruction, indicating a benefit from enforcing representation consistency.

\para{Discussion and Limitations.}
A limitation of this work is that our evaluation focuses on datasets from a single wearable device \rev{(i.e., Empatica E4)}, which allows us to isolate the core challenge of modality missingness but does not capture cross-device variability. 
While the proposed framework is modality-agnostic and does not rely on specific sensors (e.g., PPG/EDA/ACC/TEMP), validating cross-device and broader generalization remains an important direction for future work.








\vspace{-0.2cm}
\section{Conclusion}
\vspace{-0.2cm}
We presented \name, a missingness-robust pretraining framework for learning valid representations \qedit{for} multimodal wearable signals. \name utilizes a geometric orthogonal tokenizer to structurally decompose multimodal inputs into shared semantics and modality-specific residuals, which preserves information integrity while providing a rigorous inductive bias. Together with a missing-aware MoE backbone, \name enables experts to adapt to diverse modality combinations without being forced to hallucinate unobservable specific details. Overall, \name demonstrates significant advantages and strong generalization across multiple health monitoring tasks in both full-modality and modality-missing scenarios. 
\name paves the way for general-purpose pretraining under missing modalities and highlights the importance of prioritizing representation validity, which is often overlooked when learning from incomplete observations.



\newpage
\bibliography{example_paper}

\newpage
\appendix
\onecolumn
\section{Related Works}
\textbf{Wearable / Physiological Foundation Models and Self-Supervised Pretraining.}
Self-supervised pretraining on large-scale wearable signals has enabled general-purpose representations across heterogeneous sensors~\cite{abbaspourazad2024largescale,narayanswamy2025scaling,kiyasseh2021clocs}.
Beyond wearable-specific pretraining, generic time-series SSL frameworks~\cite{tonekaboniunsupervised, duan2024mf, yue2022ts2vec,nietime,goswami2024moment} have demonstrated strong label efficiency and transferability across diverse time-series domains, motivating the pursuit of general-purpose sensing representations.
While such models demonstrate broad transfer~\cite{dong2023simmtm,eldele2021time,naguiding} (e.g., imputation and downstream health/activity prediction), many formulations still train under fully observed inputs.
\rev{While recent years have seen notable progress in incomplete multimodal learning, many existing works are tailored for specific tasks (e.g., recommendation systems~\cite{lin2023contrastive} or sleep staging~\cite{shen2024robust}), designed for particular modality combinations with a dominant junction modality~\cite{dai2025babel}, or focus primarily on handling modality missingness during training~\cite{ouyang2025mmbind}. These approaches fall short of the goal of a modality-agnostic representation learning framework that can be fine-tuned for various health-related downstream tasks and effectively handle train-on-full, test-on-missing scenarios, especially in limited data settings.}
LSM-2~\cite{xu2025lsm2} with Adaptive and Inherited Masking (AIM) explicitly trains on naturally incomplete streams without pre-imputation, improving robustness to sensor dropouts. However, during training, for a missing modality that is entirely dropped, LSM-2 learns by reconstructing the raw data, which forces the model to generate unpredictable modality-specific details.

\textbf{Supervised Methods for Missing Modalities.}
Recent routing/MoE architectures Flex-MoE~\cite{yun2024flexmoe} and FuseMoE~\cite{han2024fusemoe} improve subset-robust inference by conditioning computation on observed modalities.
MAESTRO~\cite{mohapatra2025maestro} targets multimodal dynamic time series with arbitrary missing observations via symbolic tokenization, sparse cross-modal attention, and sparse MoE routing. These methods are difficult to transfer to new tasks, especially those with very limited labels~\cite{wu2024multimodal,wang2023incomplete}.

\textbf{Shared-Specific Disentanglement, Orthogonality.}
Shared-specific factorization aims to separate cross-domain semantics from domain-specific variation~\cite{liang2023factorized}, with early formulations such as DSN~\cite{bousmalis2016dsn} combining task supervision, reconstruction, and soft separation constraints.
In multimodal SSL for sensing, FOCAL~\cite{liu2023focal} factorizes each modality into shared and private embeddings and encourages orthogonality while optimizing contrastive objectives.
These methods rely on soft constraints, which fail to strictly disentangle the modalities and thus cannot prevent the model from generating inherently unpredictable modality-specific information.

\section{Downstream Dataset Details}
\begin{figure}[h]
    \centering 
    \setlength{\abovecaptionskip}{1pt}  
    \includegraphics[width=0.95\linewidth]{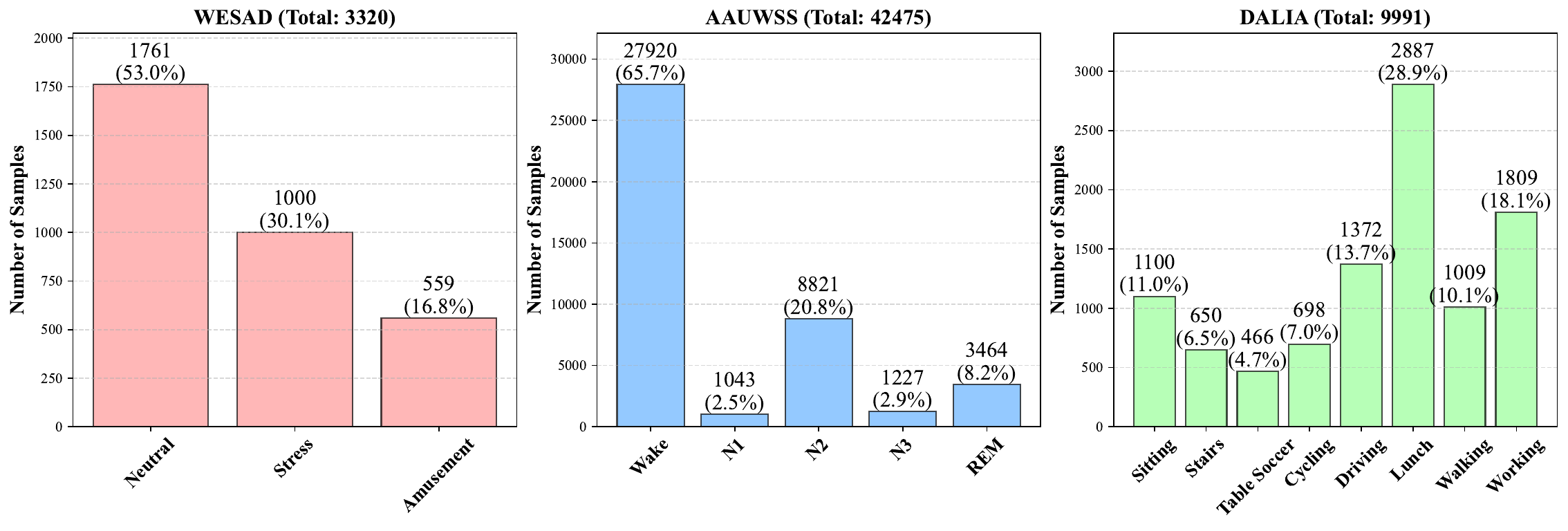}
    \caption{Distributions of Three Downstream Tasks (WESAD, AAUWSS, and DaLiA). }
    \label{fig:dataset_distribution} 
    \vspace{2cm}
\end{figure}

\clearpage
\section{\rev{Evaluation with Multiple Simultaneous Missing Modalities}}
\label{multiple_missing}

\begin{table}[!htbp]
\centering
\scriptsize
\caption{\textbf{Multiple Modality Missing Results (Macro F1).} Best results in each category (Supervised/Pretrained) are marked in \textbf{bold}, second best are \underline{underlined}.
W=WESAD, A=AAUWSS, D=DALIA.}
\label{tab:multiple_missing_f1_transposed}
\renewcommand{\arraystretch}{1.15}
\resizebox{\textwidth}{!}{
\begin{tabular}{
ll
c c c >{\columncolor{grayrow}}c
c c c c c c c >{\columncolor{grayrow}}c
}
\toprule
Missing Modality & Task & ResNet & ViT & FlexMoE & \textbf{Ours (FT)} & LSM & LSM-2 & SimCLR & MSN & CMC & MFCLR & FOCAL & \textbf{Ours (LP)} \\
\midrule

\multirow{3}{*}{PPG EDA}
& W & \underline{0.469$\pm$0.010} & \textbf{0.512$\pm$0.043} & 0.403$\pm$0.034 & 0.371$\pm$0.057 & \textbf{0.451$\pm$0.070} & \underline{0.411$\pm$0.007} & 0.393$\pm$0.025 & 0.231$\pm$0.000 & 0.318$\pm$0.031 & 0.263$\pm$0.007 & 0.268$\pm$0.049 & 0.376$\pm$0.034 \\
& A & 0.180$\pm$0.014 & 0.200$\pm$0.031 & \textbf{0.215$\pm$0.019} & \underline{0.208$\pm$0.019} & 0.182$\pm$0.022 & 0.170$\pm$0.010 & \textbf{0.216$\pm$0.023} & \underline{0.212$\pm$0.002} & 0.193$\pm$0.018 & 0.173$\pm$0.000 & 0.157$\pm$0.035 & \underline{0.212$\pm$0.015} \\
& D & 0.504$\pm$0.022 & 0.482$\pm$0.027 & \underline{0.547$\pm$0.038} & \textbf{0.720$\pm$0.022} & 0.281$\pm$0.043 & \underline{0.522$\pm$0.044} & 0.451$\pm$0.038 & 0.136$\pm$0.001 & 0.148$\pm$0.059 & 0.159$\pm$0.004 & 0.054$\pm$0.001 & \textbf{0.660$\pm$0.040} \\
\midrule

\multirow{3}{*}{PPG ACC}
& W & \textbf{0.597$\pm$0.034} & 0.529$\pm$0.045 & 0.530$\pm$0.046 & \underline{0.560$\pm$0.049} & 0.420$\pm$0.078 & \underline{0.530$\pm$0.023} & 0.209$\pm$0.081 & 0.514$\pm$0.010 & 0.420$\pm$0.080 & 0.444$\pm$0.015 & 0.352$\pm$0.174 & \textbf{0.588$\pm$0.026} \\
& A & 0.179$\pm$0.022 & 0.189$\pm$0.041 & \underline{0.211$\pm$0.030} & \textbf{0.216$\pm$0.011} & 0.176$\pm$0.004 & 0.115$\pm$0.009 & 0.169$\pm$0.006 & \underline{0.228$\pm$0.006} & 0.186$\pm$0.019 & 0.183$\pm$0.012 & 0.187$\pm$0.018 & \textbf{0.284$\pm$0.014} \\
& D & 0.216$\pm$0.016 & 0.171$\pm$0.016 & \underline{0.354$\pm$0.077} & \textbf{0.405$\pm$0.016} & 0.168$\pm$0.044 & \underline{0.401$\pm$0.019} & 0.032$\pm$0.019 & \textbf{0.408$\pm$0.001} & 0.087$\pm$0.024 & 0.302$\pm$0.002 & 0.054$\pm$0.000 & 0.399$\pm$0.020 \\
\midrule

\multirow{3}{*}{PPG TEMP}
& W & 0.466$\pm$0.080 & \underline{0.566$\pm$0.050} & 0.564$\pm$0.027 & \textbf{0.608$\pm$0.049} & 0.522$\pm$0.017 & \textbf{0.639$\pm$0.004} & 0.493$\pm$0.031 & 0.231$\pm$0.000 & 0.497$\pm$0.053 & 0.479$\pm$0.004 & 0.293$\pm$0.083 & \underline{0.582$\pm$0.049} \\
& A & 0.198$\pm$0.017 & \underline{0.228$\pm$0.018} & 0.199$\pm$0.011 & \textbf{0.229$\pm$0.020} & 0.206$\pm$0.021 & 0.217$\pm$0.002 & 0.227$\pm$0.029 & 0.190$\pm$0.010 & \underline{0.240$\pm$0.013} & 0.196$\pm$0.005 & 0.158$\pm$0.045 & \textbf{0.303$\pm$0.031} \\
& D & \underline{0.558$\pm$0.032} & 0.509$\pm$0.025 & 0.483$\pm$0.059 & \textbf{0.646$\pm$0.032} & 0.388$\pm$0.021 & \underline{0.526$\pm$0.072} & 0.400$\pm$0.047 & 0.399$\pm$0.002 & 0.189$\pm$0.079 & 0.228$\pm$0.006 & 0.064$\pm$0.021 & \textbf{0.616$\pm$0.008} \\
\midrule

\multirow{3}{*}{EDA ACC}
& W & \textbf{0.399$\pm$0.068} & 0.279$\pm$0.015 & 0.324$\pm$0.039 & \underline{0.333$\pm$0.016} & 0.335$\pm$0.057 & \textbf{0.418$\pm$0.008} & 0.211$\pm$0.083 & 0.231$\pm$0.000 & 0.340$\pm$0.074 & 0.326$\pm$0.003 & 0.266$\pm$0.089 & \underline{0.354$\pm$0.048} \\
& A & 0.158$\pm$0.029 & 0.190$\pm$0.010 & \underline{0.252$\pm$0.009} & \textbf{0.268$\pm$0.020} & 0.182$\pm$0.003 & 0.190$\pm$0.005 & 0.169$\pm$0.006 & \underline{0.228$\pm$0.001} & \textbf{0.245$\pm$0.019} & 0.175$\pm$0.001 & 0.181$\pm$0.002 & 0.204$\pm$0.013 \\
& D & 0.156$\pm$0.013 & 0.128$\pm$0.027 & \underline{0.173$\pm$0.029} & \textbf{0.352$\pm$0.013} & 0.084$\pm$0.016 & 0.174$\pm$0.033 & 0.044$\pm$0.039 & 0.148$\pm$0.002 & 0.123$\pm$0.036 & 0.142$\pm$0.003 & \underline{0.193$\pm$0.057} & \textbf{0.306$\pm$0.025} \\
\midrule

\multirow{3}{*}{EDA TEMP}
& W & \underline{0.435$\pm$0.019} & 0.411$\pm$0.033 & 0.425$\pm$0.035 & \textbf{0.464$\pm$0.033} & 0.413$\pm$0.019 & 0.264$\pm$0.022 & 0.383$\pm$0.021 & 0.231$\pm$0.000 & 0.255$\pm$0.033 & 0.325$\pm$0.007 & \underline{0.432$\pm$0.043} & \textbf{0.460$\pm$0.024} \\
& A & 0.161$\pm$0.019 & \underline{0.201$\pm$0.015} & \underline{0.201$\pm$0.015} & \textbf{0.247$\pm$0.010} & 0.225$\pm$0.008 & 0.173$\pm$0.002 & 0.224$\pm$0.010 & 0.173$\pm$0.001 & 0.150$\pm$0.018 & 0.178$\pm$0.000 & \underline{0.226$\pm$0.003} & \textbf{0.230$\pm$0.016} \\
& D & \underline{0.303$\pm$0.016} & 0.259$\pm$0.021 & 0.274$\pm$0.103 & \textbf{0.561$\pm$0.024} & 0.223$\pm$0.015 & 0.252$\pm$0.042 & 0.298$\pm$0.053 & 0.095$\pm$0.001 & 0.263$\pm$0.062 & 0.090$\pm$0.001 & \underline{0.463$\pm$0.022} & \textbf{0.493$\pm$0.029} \\
\midrule

\multirow{3}{*}{ACC TEMP}
& W & 0.503$\pm$0.066 & 0.505$\pm$0.005 & \underline{0.514$\pm$0.014} & \textbf{0.563$\pm$0.017} & 0.441$\pm$0.062 & \underline{0.549$\pm$0.066} & 0.242$\pm$0.137 & 0.231$\pm$0.000 & 0.487$\pm$0.066 & 0.490$\pm$0.003 & \textbf{0.561$\pm$0.054} & \textbf{0.561$\pm$0.056} \\
& A & 0.134$\pm$0.049 & 0.165$\pm$0.015 & \underline{0.193$\pm$0.024} & \textbf{0.277$\pm$0.026} & 0.235$\pm$0.021 & 0.230$\pm$0.002 & 0.165$\pm$0.017 & 0.183$\pm$0.000 & \underline{0.258$\pm$0.015} & 0.225$\pm$0.002 & 0.194$\pm$0.005 & \textbf{0.271$\pm$0.014} \\
& D & 0.102$\pm$0.021 & \underline{0.187$\pm$0.038} & 0.181$\pm$0.018 & \textbf{0.437$\pm$0.025} & 0.237$\pm$0.016 & 0.241$\pm$0.007 & 0.024$\pm$0.015 & 0.232$\pm$0.001 & 0.222$\pm$0.030 & \underline{0.277$\pm$0.004} & 0.199$\pm$0.022 & \textbf{0.387$\pm$0.008} \\
\midrule

\multirow{3}{*}{PPG EDA ACC}
& W & \textbf{0.408$\pm$0.077} & 0.270$\pm$0.000 & \underline{0.312$\pm$0.052} & 0.279$\pm$0.039 & 0.225$\pm$0.012 & \textbf{0.448$\pm$0.019} & 0.211$\pm$0.083 & 0.231$\pm$0.000 & 0.299$\pm$0.069 & 0.231$\pm$0.000 & 0.232$\pm$0.004 & \underline{0.303$\pm$0.044} \\
& A & 0.172$\pm$0.019 & 0.185$\pm$0.009 & \underline{0.220$\pm$0.005} & \textbf{0.231$\pm$0.021} & 0.173$\pm$0.000 & \underline{0.185$\pm$0.000} & 0.168$\pm$0.006 & 0.183$\pm$0.003 & 0.180$\pm$0.014 & 0.173$\pm$0.000 & 0.177$\pm$0.007 & \textbf{0.215$\pm$0.003} \\
& D & 0.130$\pm$0.018 & 0.062$\pm$0.009 & \underline{0.155$\pm$0.035} & \textbf{0.213$\pm$0.018} & 0.054$\pm$0.000 & 0.158$\pm$0.037 & 0.026$\pm$0.014 & 0.098$\pm$0.000 & 0.053$\pm$0.007 & \underline{0.181$\pm$0.007} & 0.054$\pm$0.000 & \textbf{0.183$\pm$0.020} \\
\midrule

\multirow{3}{*}{PPG EDA TEMP}
& W & \underline{0.439$\pm$0.023} & 0.414$\pm$0.045 & 0.418$\pm$0.068 & \textbf{0.471$\pm$0.028} & 0.399$\pm$0.066 & \underline{0.421$\pm$0.015} & 0.383$\pm$0.019 & 0.231$\pm$0.000 & 0.250$\pm$0.050 & 0.258$\pm$0.003 & 0.239$\pm$0.011 & \textbf{0.423$\pm$0.020} \\
& A & 0.159$\pm$0.014 & \textbf{0.193$\pm$0.013} & 0.153$\pm$0.029 & \underline{0.190$\pm$0.020} & 0.160$\pm$0.026 & \underline{0.222$\pm$0.008} & \textbf{0.226$\pm$0.009} & 0.192$\pm$0.001 & 0.171$\pm$0.006 & 0.173$\pm$0.000 & 0.149$\pm$0.049 & 0.180$\pm$0.022 \\
& D & \underline{0.307$\pm$0.042} & 0.282$\pm$0.023 & 0.268$\pm$0.072 & \textbf{0.531$\pm$0.020} & 0.199$\pm$0.018 & 0.258$\pm$0.053 & \underline{0.300$\pm$0.053} & 0.201$\pm$0.004 & 0.103$\pm$0.064 & 0.065$\pm$0.002 & 0.054$\pm$0.000 & \textbf{0.458$\pm$0.034} \\
\midrule

\multirow{3}{*}{PPG ACC TEMP}
& W & 0.491$\pm$0.088 & 0.501$\pm$0.040 & \underline{0.520$\pm$0.022} & \textbf{0.567$\pm$0.033} & 0.292$\pm$0.080 & \textbf{0.646$\pm$0.045} & 0.242$\pm$0.137 & 0.231$\pm$0.000 & 0.382$\pm$0.091 & 0.415$\pm$0.016 & 0.314$\pm$0.166 & \underline{0.590$\pm$0.023} \\
& A & 0.150$\pm$0.054 & 0.140$\pm$0.046 & \underline{0.178$\pm$0.021} & \textbf{0.231$\pm$0.013} & 0.173$\pm$0.000 & 0.151$\pm$0.000 & 0.165$\pm$0.017 & \underline{0.270$\pm$0.001} & 0.203$\pm$0.026 & 0.199$\pm$0.016 & 0.180$\pm$0.013 & \textbf{0.304$\pm$0.028} \\
& D & 0.072$\pm$0.007 & 0.096$\pm$0.029 & \underline{0.165$\pm$0.027} & \textbf{0.262$\pm$0.009} & 0.167$\pm$0.047 & \underline{0.261$\pm$0.023} & 0.024$\pm$0.015 & 0.165$\pm$0.000 & 0.039$\pm$0.022 & 0.228$\pm$0.002 & 0.054$\pm$0.000 & \textbf{0.268$\pm$0.013} \\
\midrule

\multirow{3}{*}{EDA ACC TEMP}
& W & 0.151$\pm$0.047 & 0.196$\pm$0.040 & \underline{0.257$\pm$0.049} & \textbf{0.391$\pm$0.030} & 0.304$\pm$0.044 & 0.260$\pm$0.007 & 0.173$\pm$0.052 & 0.231$\pm$0.000 & 0.216$\pm$0.030 & \underline{0.308$\pm$0.006} & 0.253$\pm$0.038 & \textbf{0.364$\pm$0.035} \\
& A & 0.119$\pm$0.050 & \underline{0.180$\pm$0.003} & 0.167$\pm$0.032 & \textbf{0.242$\pm$0.006} & 0.182$\pm$0.004 & 0.157$\pm$0.001 & 0.173$\pm$0.000 & 0.179$\pm$0.001 & \textbf{0.230$\pm$0.018} & 0.176$\pm$0.001 & 0.178$\pm$0.001 & \underline{0.214$\pm$0.011} \\
& D & 0.037$\pm$0.010 & 0.066$\pm$0.011 & \underline{0.090$\pm$0.024} & \textbf{0.311$\pm$0.030} & 0.082$\pm$0.010 & 0.066$\pm$0.011 & 0.024$\pm$0.015 & 0.039$\pm$0.000 & 0.083$\pm$0.015 & 0.092$\pm$0.001 & \underline{0.133$\pm$0.051} & \textbf{0.230$\pm$0.035} \\
\bottomrule
\end{tabular}
}
\end{table}

\begin{table}[!htbp]
\vspace{-0.2cm}
\centering
\scriptsize
\caption{\textbf{WEEE Modality Missing Results (MAE $\downarrow$).} Best results in each category (Supervised/Pretrained) are marked in \textbf{bold}, second best are \underline{underlined}.
To further validate generalizability beyond classification, we introduced a new regression benchmark, WEEE (predicting $\text{VO}_2$), evaluated under all 10 possible missing patterns ($C_{4}^{2} + C_{4}^{3}$).
VCR achieves strong performance.}
\label{tab:full_missing_modalities_mae_transposed}
\renewcommand{\arraystretch}{1.3}
\setlength{\aboverulesep}{0pt}
\setlength{\belowrulesep}{0pt}
\resizebox{\textwidth}{!}{
\begin{tabular}{
l
c c c >{\columncolor{grayrow}}c !{\vrule width 2pt}
c c c c c c c >{\columncolor{grayrow}}c
}
\toprule
Missing Modality & ResNet & ViT & FlexMoE & \textbf{Ours (FT)} & LSM & LSM-2 & SimCLR & MSN & CMC & MFCLR & FOCAL & \textbf{Ours (LP)} \\
\midrule
PPG EDA & 5.782$\pm$0.696 & 6.451$\pm$1.257 & \underline{4.038$\pm$0.600} & \textbf{3.518$\pm$0.232} & 6.654$\pm$0.126 & 7.838$\pm$0.062 & 6.196$\pm$0.326 & 6.728$\pm$0.026 & 11.839$\pm$6.394 & \underline{4.938$\pm$0.085} & 24.576$\pm$4.157 & \textbf{3.756$\pm$0.071} \\
PPG ACC & 4.760$\pm$0.403 & 5.102$\pm$1.228 & \underline{4.368$\pm$0.463} & \textbf{4.028$\pm$0.335} & \textbf{3.907$\pm$0.079} & 6.246$\pm$0.072 & 19.697$\pm$13.521 & 5.036$\pm$0.016 & 6.115$\pm$1.836 & \underline{4.249$\pm$0.146} & 26.490$\pm$3.227 & 4.448$\pm$0.090 \\
PPG TEMP & \underline{3.627$\pm$0.261} & 3.878$\pm$0.283 & 3.781$\pm$0.353 & \textbf{3.278$\pm$0.377} & \underline{3.810$\pm$0.094} & 6.031$\pm$0.081 & 4.105$\pm$0.134 & 4.392$\pm$0.003 & 5.159$\pm$0.787 & 4.905$\pm$0.146 & 26.953$\pm$2.343 & \textbf{3.382$\pm$0.065} \\
EDA ACC & 5.125$\pm$0.553 & 3.955$\pm$0.149 & \underline{3.106$\pm$0.318} & \textbf{2.985$\pm$0.134} & 5.263$\pm$0.300 & \underline{4.658$\pm$0.012} & 20.444$\pm$13.550 & 17.254$\pm$0.118 & 7.631$\pm$3.671 & 5.524$\pm$0.067 & 6.613$\pm$2.463 & \textbf{3.621$\pm$0.062} \\
EDA TEMP & 4.494$\pm$0.252 & 6.112$\pm$0.610 & \underline{3.529$\pm$0.350} & \textbf{3.072$\pm$0.167} & 5.945$\pm$0.159 & 5.090$\pm$0.018 & 7.243$\pm$0.194 & 5.598$\pm$0.014 & 9.957$\pm$3.296 & 6.153$\pm$0.105 & \underline{4.926$\pm$1.251} & \textbf{2.973$\pm$0.082} \\
ACC TEMP & 4.260$\pm$0.441 & 4.391$\pm$0.717 & \underline{2.958$\pm$0.228} & \textbf{2.899$\pm$0.134} & 3.741$\pm$0.191 & \underline{3.678$\pm$0.019} & 21.312$\pm$14.262 & 4.018$\pm$0.005 & 6.205$\pm$0.657 & 4.782$\pm$0.054 & 3.754$\pm$0.751 & \textbf{3.421$\pm$0.035} \\
PPG EDA ACC & 5.988$\pm$0.419 & 5.046$\pm$0.256 & \textbf{4.200$\pm$0.631} & \underline{4.545$\pm$0.213} & 5.352$\pm$0.290 & 6.917$\pm$0.076 & 20.536$\pm$13.583 & 11.968$\pm$0.084 & 9.373$\pm$5.386 & \textbf{4.761$\pm$0.078} & 24.825$\pm$6.030 & \underline{4.764$\pm$0.076} \\
PPG EDA TEMP & \underline{4.218$\pm$0.240} & 6.509$\pm$0.702 & 4.671$\pm$0.064 & \textbf{4.081$\pm$0.366} & 6.892$\pm$0.204 & 9.449$\pm$0.057 & 7.225$\pm$0.205 & 6.111$\pm$0.019 & 11.641$\pm$5.763 & \underline{5.532$\pm$0.166} & 25.288$\pm$3.739 & \textbf{4.178$\pm$0.133} \\
PPG ACC TEMP & \underline{4.495$\pm$0.740} & 5.232$\pm$1.095 & \textbf{4.205$\pm$0.503} & 5.058$\pm$0.378 & \textbf{3.815$\pm$0.122} & 14.385$\pm$0.310 & 21.308$\pm$14.218 & \underline{4.322$\pm$0.007} & 6.619$\pm$2.190 & 4.861$\pm$0.173 & 27.203$\pm$2.115 & 4.642$\pm$0.042 \\
EDA ACC TEMP & 4.938$\pm$0.468 & 5.559$\pm$0.352 & \underline{3.623$\pm$0.417} & \textbf{2.982$\pm$0.171} & 5.095$\pm$0.222 & \underline{4.929$\pm$0.010} & 22.162$\pm$14.438 & 6.890$\pm$0.036 & 7.477$\pm$2.819 & 6.245$\pm$0.108 & 5.587$\pm$2.168 & \textbf{3.596$\pm$0.045} \\
\bottomrule
\end{tabular}
}
\end{table}

\begin{table}[!htbp]
\centering
\scriptsize
\caption{\textbf{WEEE Modality Missing Results (R$^2$ $\uparrow$).} Best results in each category (Supervised/Pretrained) are marked in \textbf{bold}, second best are \underline{underlined}.
To further validate generalizability beyond classification, we introduced a new regression benchmark, WEEE (predicting $\text{VO}_2$), evaluated under all 10 possible missing patterns ($C_{4}^{2} + C_{4}^{3}$).
VCR achieves strong performance.}
\label{tab:full_missing_modalities_r2_transposed}
\renewcommand{\arraystretch}{1.3}
\setlength{\aboverulesep}{0pt}
\setlength{\belowrulesep}{0pt}
\resizebox{\textwidth}{!}{
\begin{tabular}{
l
c c c >{\columncolor{grayrow}}c !{\vrule width 2pt}
c c c c c c c >{\columncolor{grayrow}}c
}
\toprule
Missing Modality & ResNet & ViT & FlexMoE & \textbf{Ours (FT)} & LSM & LSM-2 & SimCLR & MSN & CMC & MFCLR & FOCAL & \textbf{Ours (LP)} \\
\midrule
PPG EDA & -0.538$\pm$0.339 & -0.892$\pm$0.699 & \underline{0.098$\pm$0.258} & \textbf{0.423$\pm$0.064} & -0.581$\pm$0.058 & -1.139$\pm$0.037 & -0.487$\pm$0.167 & -0.734$\pm$0.015 & -4.939$\pm$5.727 & \underline{0.042$\pm$0.028} & -18.182$\pm$5.780 & \textbf{0.240$\pm$0.078} \\
PPG ACC & -0.214$\pm$0.146 & -0.353$\pm$0.510 & \underline{0.119$\pm$0.132} & \textbf{0.222$\pm$0.131} & \textbf{0.348$\pm$0.047} & -0.578$\pm$0.036 & -16.357$\pm$19.761 & -0.282$\pm$0.012 & -0.690$\pm$0.959 & \underline{0.278$\pm$0.044} & -20.681$\pm$4.910 & 0.144$\pm$0.038 \\
PPG TEMP & \underline{0.377$\pm$0.060} & 0.289$\pm$0.117 & 0.300$\pm$0.080 & \textbf{0.487$\pm$0.104} & \underline{0.354$\pm$0.030} & -0.512$\pm$0.036 & 0.223$\pm$0.058 & 0.237$\pm$0.001 & -0.148$\pm$0.276 & 0.036$\pm$0.060 & -21.087$\pm$3.787 & \textbf{0.432$\pm$0.037} \\
EDA ACC & -0.250$\pm$0.280 & 0.245$\pm$0.069 & \underline{0.492$\pm$0.083} & \textbf{0.597$\pm$0.026} & -0.057$\pm$0.101 & \underline{0.018$\pm$0.003} & -17.426$\pm$20.684 & -8.536$\pm$0.123 & -1.581$\pm$2.484 & -0.221$\pm$0.037 & -0.925$\pm$1.208 & \textbf{0.443$\pm$0.022} \\
EDA TEMP & 0.134$\pm$0.117 & -0.548$\pm$0.307 & \underline{0.356$\pm$0.110} & \textbf{0.547$\pm$0.046} & -0.364$\pm$0.064 & -0.071$\pm$0.005 & -1.007$\pm$0.131 & -0.155$\pm$0.005 & -2.813$\pm$2.405 & -0.501$\pm$0.063 & \underline{-0.047$\pm$0.472} & \textbf{0.597$\pm$0.027} \\
ACC TEMP & 0.073$\pm$0.231 & 0.059$\pm$0.234 & \underline{0.539$\pm$0.056} & \textbf{0.576$\pm$0.028} & 0.378$\pm$0.060 & \underline{0.403$\pm$0.005} & -18.945$\pm$21.854 & 0.303$\pm$0.002 & -0.694$\pm$0.343 & -0.000$\pm$0.023 & 0.380$\pm$0.241 & \textbf{0.529$\pm$0.011} \\
PPG EDA ACC & -0.842$\pm$0.205 & -0.316$\pm$0.177 & \textbf{0.210$\pm$0.206} & \underline{0.161$\pm$0.073} & -0.091$\pm$0.096 & -0.889$\pm$0.038 & -17.565$\pm$20.836 & -3.929$\pm$0.061 & -3.061$\pm$4.343 & \textbf{0.127$\pm$0.013} & -19.198$\pm$8.577 & \underline{-0.021$\pm$0.037} \\
PPG EDA TEMP & \textbf{0.227$\pm$0.099} & -0.730$\pm$0.391 & -0.059$\pm$0.082 & \underline{0.184$\pm$0.172} & -0.678$\pm$0.096 & -2.200$\pm$0.042 & -0.995$\pm$0.136 & -0.373$\pm$0.009 & -4.546$\pm$4.721 & \underline{-0.182$\pm$0.071} & -19.038$\pm$5.364 & \textbf{0.028$\pm$0.124} \\
PPG ACC TEMP & \underline{-0.050$\pm$0.355} & -0.326$\pm$0.406 & \textbf{0.250$\pm$0.159} & -0.115$\pm$0.149 & \textbf{0.344$\pm$0.078} & -7.101$\pm$0.308 & -18.911$\pm$21.788 & \underline{0.239$\pm$0.003} & -0.886$\pm$1.086 & 0.080$\pm$0.064 & -21.557$\pm$3.426 & 0.078$\pm$0.027 \\
EDA ACC TEMP & -0.124$\pm$0.253 & -0.269$\pm$0.187 & \underline{0.310$\pm$0.146} & \textbf{0.612$\pm$0.031} & \underline{-0.012$\pm$0.062} & -0.196$\pm$0.003 & -20.357$\pm$23.197 & -0.834$\pm$0.020 & -1.327$\pm$1.762 & -0.539$\pm$0.062 & -0.421$\pm$0.973 & \textbf{0.428$\pm$0.019} \\
\bottomrule
\end{tabular}
}
\end{table}

\section{\rev{Ablation on Shared/Specific Features.}}
To further validate the necessity of preserving both modality-shared and modality-specific information, we conduct an ablation study on the feature types generated by our orthogonal tokenizer. As shown in Table~\ref{tab:token_ablation}, we evaluate the representation quality directly under the full-modality setting using different subsets of the tokens: only the shared components ($\mathbf{s}$\_only), only the specific residuals ($\mathbf{p}$\_only), and the combination of both (both ($\mathbf{s}$+$\mathbf{p}$)). 

The results demonstrate that utilizing both components consistently yields the best Macro-F1 performance across three downstream health tasks (WESAD, AAUWSS, and DALIA). Dropping either the modal-shared or the modality-specific tokens leads to notable performance degradation. This confirms that these two part of features are both crucial for accurate human state estimation.
\begin{table*}[ht]
\centering
\renewcommand{\arraystretch}{1.15}
\caption{\textbf{Ablation Study on Feature Types (Macro F1 $\uparrow$).} Macro-F1 results evaluated directly under the full modality setting, using shared ($\mathbf{s}$) and specific ($\mathbf{p}$) tokens generated by the tokenizer.}
\label{tab:token_ablation}
\begin{tabular}{l c c c}
\toprule
Ablation Mode & WESAD & AAUWSS & DALIA \\
\midrule
both ($\mathbf{s}$+$\mathbf{p}$) & \textbf{0.6593$\pm$0.0152} & \textbf{0.3178$\pm$0.0094} & \textbf{0.7425$\pm$0.0105} \\
$\mathbf{s}$\_only & 0.5842$\pm$0.0183 & 0.2641$\pm$0.0112 & 0.7176$\pm$0.0132 \\
$\mathbf{p}$\_only & 0.5400$\pm$0.0215 & 0.2968$\pm$0.0145 & 0.6988$\pm$0.0167 \\
\bottomrule
\end{tabular}
\end{table*}

\section{Ablation on $\lambda_{recon}/\lambda_{align}$:}
\begin{table}[h]
\centering
\caption{Analysis of the weight ratio $\lambda_{recon} / \lambda_{align}$ in the Tokenizer loss.}
\label{tab:lambda_ablation}
\begin{tabular}{lccc}
\toprule
Ratio & WESAD ($F_1$) & AAUWSS ($F_1$) & DaLiA ($F_1$) \\
\midrule
0.1 & $0.640 \pm 0.042$ & $0.287 \pm 0.006$ & $0.752 \pm 0.003$ \\
1.0 & $0.583 \pm 0.035$ & $0.275 \pm 0.009$ & $0.662 \pm 0.007$ \\
10.0 & $0.623 \pm 0.048$ & $0.285 \pm 0.007$ & $0.685 \pm 0.005$ \\
\bottomrule
\end{tabular}
\end{table}
Given the disparate scales and optimization difficulties of the reconstruction and contrastive losses, we evaluate various $\lambda_{recon}/\lambda_{align}$ ratios, as reported in Table~\ref{tab:lambda_ablation}. We observe that optimal performance is achieved at a ratio of 0.1. An excessively large $\lambda_{recon}$ incentivizes the model to offload the majority of information into the modality-specific residual $\mathbf{p}$ to minimize reconstruction error. Consequently, the shared representation $\mathbf{s}$ degenerates into modality-consistent noise rather than capturing meaningful semantics.

\section{Ablation on MoE Block Placement}
\begin{table}[h]
\centering
\caption{Impact of MoE layer index in the backbone.}
\label{tab:moe_position_ablation}
\begin{tabular}{lccc}
\toprule
Layer & WESAD ($F_1$) & AAUWSS ($F_1$) & DaLiA ($F_1$) \\
\midrule
Layer 0 & $0.602 \pm 0.033$ & $0.236 \pm 0.005$ & $0.711 \pm 0.008$ \\
Layer 2 & $0.640 \pm 0.042$ & $0.287 \pm 0.006$ & $0.752 \pm 0.003$ \\
Layer 4 & $0.623 \pm 0.046$ & $0.249 \pm 0.008$ & $0.730 \pm 0.006$ \\
\bottomrule
\end{tabular}
\end{table}
We evaluate the impact of MoE layer placement within the missing-aware backbone and find that Layer 2 (0-indexed) yields optimal performance, as shown in Table~\ref{tab:moe_position_ablation}. We attribute this to a trade-off between contextual awareness and specialization: if the MoE layer is placed too early, tokens lack sufficient self-attention depth to perceive global modality availability, rendering routing decisions susceptible to initialization noise. Conversely, if placed too late, cross-modal representations become excessively fused, diminishing the experts' ability to specialize in handling distinct modality combinations.

\section{Ablation on Shared Subspace Rank}
\begin{figure}[h]
    \centering 
    \setlength{\abovecaptionskip}{1pt}  
    \includegraphics[width=0.5\linewidth]{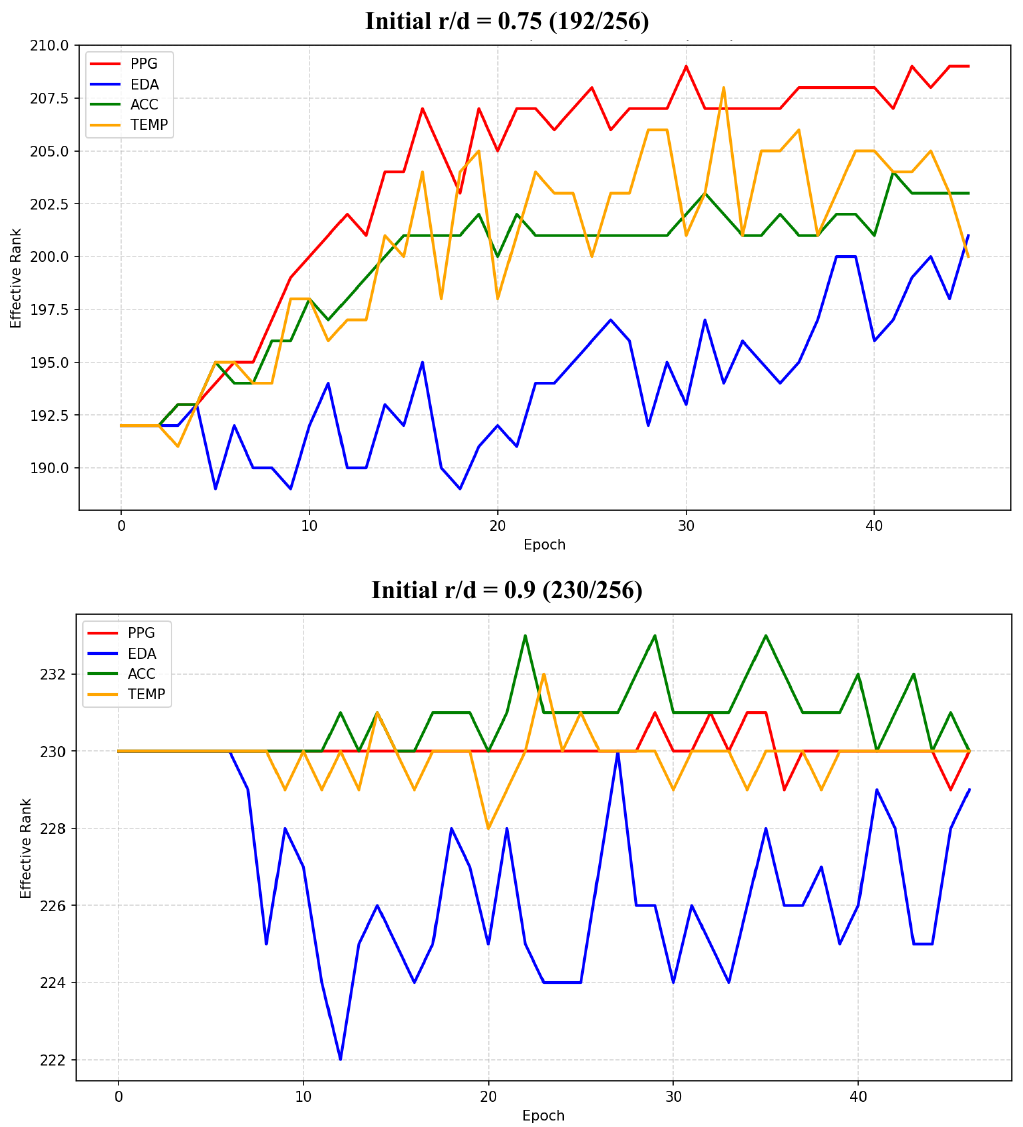}
    \caption{Rank evolution during training. }
    \label{fig:rank_ratio} 
    \vspace{-0.6cm}
\end{figure}

In our framework, the rank $r$ governs the allocation of information capacity between the shared and specific subspaces. To regulate this, we employ a gating mechanism that determines the activation of each orthogonal basis vector within the projection operator $\mathcal{P}$, thereby adaptively learning its effective rank. We evaluate various initial rank-to-dimension ratios ($r/d$) and observe that an initialization of 0.9 yields optimal performance, as shown in Table~\ref{tab:rank_ablation}. Furthermore, visualizing the rank evolution for initializations of 0.75 and 0.9 in Figure~\ref{fig:rank_ratio} reveals a consistent trend: the effective rank of $\mathcal{P}$ across all modalities converges to a value exceeding 0.8 during training.
\begin{table}[htbp]
\centering
\caption{Ablation study on the initial ratio of shared subspace rank $r/d$.}
\label{tab:rank_ablation}
\begin{tabular}{lccc}
\toprule
Init Ratio & WESAD ($F_1$) & AAUWSS ($F_1$) & DaLiA ($F_1$) \\
\midrule
0.5 & $0.594 \pm 0.038$ & $0.251 \pm 0.008$ & $0.668 \pm 0.006$ \\
0.75 & $0.644 \pm 0.045$ & $0.282 \pm 0.005$ & $0.671 \pm 0.005$ \\
0.9 & $0.640 \pm 0.042$ & $0.287 \pm 0.006$ & $0.752 \pm 0.003$ \\
\bottomrule
\end{tabular}
\vspace{-0.5cm}
\end{table}

\section{Hyperparameter Settings}
\label{appendix:hyperparameter}
\rev{For the loss weights tuning, we fix $\lambda_{align}$ = 1 and $\lambda_{rand}$ = 1, so the main coefficients are: $\lambda_{recon}$, $\lambda_{white}$, and $\lambda_{struct}$. 
For Tokenizer, $\lambda_{recon}$ / $\lambda_{align}$ controls the trade-off between reconstructability and cross-modal shared semantics. If $\lambda_{recon}$ is too large, the model can offload part of the shared information into the specific branch $\textbf{p}$, because reconstructing from $\textbf{p}$ is easier to optimize than aligning the shared branch $\textbf{s}$ across modalities. This weakens the semantic content of $\textbf{s}$.
This behavior is consistent with our ablation in Appendix Table~\ref{tab:lambda_ablation}, where: $\lambda_{recon}$ / $\lambda_{align}$ = 0.1 performs best across all three downstream tasks.
We keep $\lambda_{white}$ small (0.05) because it serves as a regularizer for approximate whitening/independence rather than a primary task objective, and thus should not dominate semantic learning.
For Backbone, $\mathcal{L}_{rand}$ and $\mathcal{L}_{struct}$ are both latent-space reconstruction losses with matched form, so their scales are much more comparable than the tokenizer objectives. They play complementary roles: $\mathcal{L}_{rand}$ learns general contextual semantics
$\mathcal{L}_{struct}$ explicitly trains robustness to modality-level missingness
Since both are essential in our setting, we use the balanced default: $\lambda_{rand}$ : $\lambda_{struct}$ = 1:1.}

\begin{table}[ht]
\centering
\caption{Hyperparameters for Orthogonal Tokenizer pre-training.}
\label{tab:tokenizer_params}
\begin{tabular}{lcc}
\toprule
\multicolumn{2}{c}{\textbf{Hyperparameters}} & \textbf{Values} \\
\midrule
\multirow{4}{*}{Input Configuration} 
 & Sampling Rates (Hz) & \{PPG:64, ACC:32, EDA:4, TEMP:4\} \\
 & Input Channels & \{1, 3, 1, 1\} \\
 & Window Duration & 10s \\
 & Patch Size & 1s \\
\midrule
\multirow{4}{*}{Encoder \& Projector} 
 & Encoder Architecture & ResNet-1D \\
 & Embedding Dimension ($d_{model}$) & 256 \\
 & Initial Rank Ratio ($r/d$) & 0.9 \\

\midrule
\multirow{7}{*}{Optimization} 
 & Batch Size & 256 \\
 & Optimizer & AdamW \\
 & Learning Rate & $1 \times 10^{-4}$ \\
 & Weight Decay & $1 \times 10^{-5}$ \\
 & Warmup Epochs & 10 \\
 & Reconstruction Weight ($\lambda_{recon}$) & 0.1 \\
 & Whitening Weight ($\lambda_{white}$) & 0.05 \\
\bottomrule
\end{tabular}
\end{table}

\begin{table}[ht]
\centering
\caption{Hyperparameters for Missing-Aware MoE Backbone pre-training.}
\label{tab:backbone_params}
\begin{tabular}{lcc}
\toprule
\multicolumn{2}{c}{\textbf{Hyperparameters}} & \textbf{Values} \\
\midrule
\multirow{6}{*}{Transformer Architecture} 
 & Encoder Layers & 6 \\
 & Decoder Layers & 2 \\
 & Hidden Size ($d_{model}$) & 512 \\
 & Attention Heads & 8 \\
 & Head Dimension & 64 \\
 & Dropout & 0.1 \\
\midrule
\multirow{4}{*}{Mixture of Experts (MoE)} 
 & MoE Layer Index & 2 (0-indexed) \\
 & Number of Experts & 8 \\
 & Top-$k$ Routing & 2 \\
 & Capacity Factor & 1.25 \\
\midrule
\multirow{7}{*}{Optimization \& Masking} 
 & Masking Ratio & 0.75 \\
 & Batch Size & 1024 \\
 & Optimizer & AdamW \\
 & Learning Rate & $1 \times 10^{-4}$ \\
 & Weight Decay & 0.05 \\
 & Adam $\beta$ & (0.9, 0.999) \\
 & Warmup Epochs & 10 \\
\bottomrule
\end{tabular}
\end{table}

\clearpage
\rev{\section{Compute Resources}
All experiments, including pretraining and downstream evaluations, are conducted on a NVIDIA RTX 5090 GPU (32GB VRAM), an Intel(R) Xeon(R) Platinum 8470Q CPU (25 vCPUs), and 90GB of system memory. The complete pretraining pipeline of our VCR framework requires approximately 11.67 GPU-hours (including both the orthogonal tokenizer and the missing-aware MoE backbone). Taking into account the evaluation of baselines, cross-validation on multiple downstream tasks, and failed experiments, the total computational budget of this research project is estimated to be approximately 2400 GPU hours.}

\rev{\section{Broader Impacts}
\textbf{Positive Societal Impacts}: The primary positive impact of \name lies in advancing continuous, non-invasive health monitoring. By significantly improving model robustness against real-world modality missingness (e.g., sensor failure, user non-compliance, or hardware limitations), \name makes intelligent wearable health applications more reliable and accessible. This can potentially lower the barrier to entry for digital health technologies, allowing users with cheaper or partial sensor configurations to still benefit from high-quality physiological tracking. Furthermore, by relying on self-supervised learning, \name reduces the reliance on large-scale expert annotations, which are notoriously expensive and time-consuming to obtain in the medical and healthcare domains.}

\rev{\textbf{Potential Negative Impacts and Risks}: Wearable signals can infer highly sensitive personal states. If misused, such robust tracking systems could be deployed for continuous surveillance by employers or insurance companies without the user's explicit consent, potentially leading to privacy violations.}

\rev{\section{License}
The license information for multiple datasets: BIG IDEAs (ODC-By 1.0), UE4W (CC-BY 4.0), CAN-STRESS (CC-BY 4.0), AAUWSS (CC-BY 4.0), WEEE (CC-BY 4.0), DaLiA (CC-BY 4.0) and WESAD (CC-BY 4.0). }


\end{document}